\documentclass[letterpaper]{article} 
\usepackage{aaai2026}  
\usepackage{times}  
\usepackage{helvet}  
\usepackage{courier}  
\usepackage{graphicx} 
\usepackage{amsmath}
\usepackage{xcolor}
\usepackage{amsfonts}
\usepackage{natbib}  
\usepackage{caption} 
\usepackage{algorithm}
\usepackage{algorithmic}

\usepackage{newfloat}
\usepackage{listings}
\usepackage{booktabs}       
\usepackage{tcolorbox}
\usepackage{longtable} 
\usepackage{makecell}
\usepackage{colortbl}
\DeclareCaptionStyle{ruled}{labelfont=normalfont,labelsep=colon,strut=off} 
\floatstyle{ruled}
\newfloat{listing}{tb}{lst}{}
\floatname{listing}{Listing}
\newcommand{\benchmark}{\textsc{m$^3$TQA}}
\newcommand{\instruct}{\textsc{m$^3$TQA-Instruct}}

\title{\benchmark{}: Massively Multilingual Multitask Table Question Answering}
\author{
    Daixin Shu\textsuperscript{\rm 1}, 
    Jian Yang\textsuperscript{\rm 1}\thanks{\ \ Corresponding Author.}, 
    {\bf Zhenhe Wu}\textsuperscript{\rm 1}, 
    {\bf Xianjie Wu}\textsuperscript{\rm 1}, 
    {\bf Xianfu Cheng}\textsuperscript{\rm 1},
    {\bf Xiangyuan Guan}\textsuperscript{\rm 1}, \\
    {\bf Yanghai Wang}\textsuperscript{\rm 1},
    {\bf Pengfei Wu}\textsuperscript{\rm 3}, 
    {\bf Tingyang Yang}\textsuperscript{\rm 1}, 
    {\bf Hualei Zhu}\textsuperscript{\rm 1},  
    {\bf Wei Zhang}\textsuperscript{\rm 1},  \\
    {\bf Ge Zhang}\textsuperscript{\rm 2}, 
    {\bf Jiaheng Liu}\textsuperscript{\rm 3},
    {\bf Zhoujun Li}\textsuperscript{\rm 1} \\
    \textsuperscript{\rm 1}CCSE, Beihang University,~\textsuperscript{\rm 2}M-A-P,~\textsuperscript{\rm 3}Nanjing University\\
    \texttt{\{shudx,jiayang\}@buaa.edu.cn} \\
}
\usepackage{bibentry}

\begin{document}

\maketitle

\begin{abstract}
Tabular data is a fundamental component of real-world information systems, yet most research in table understanding remains confined to English, leaving multilingual comprehension significantly underexplored. Existing multilingual table benchmarks suffer from geolinguistic imbalance—overrepresenting Indo-European and high-resource languages—and insufficient scale for rigorous cross-lingual analysis. To address these limitations, we introduce a complete framework to enhance the massively multilingual multitask table question answering by introducing the instruction and reinforcement learning dataset \instruct{}, a large-scale benchmark spanning 97 languages across diverse language families, including underrepresented and low-resource ones. Further, we construct \benchmark{} by curating 50 real-world tables in Chinese and English, then applying a robust, six-step LLM-based translation pipeline powered by DeepSeek and GPT-4o, achieving high translation fidelity (median BLEU: 60.19) as validated by back-translation. The benchmark includes 2,916 professionally annotated question-answering pairs across four tasks designed to evaluate nuanced table reasoning capabilities. Experiments on state-of-the-art LLMs reveal critical insights into cross-lingual generalization, showing that synthetically generated, unannotated QA data can significantly boost performance—especially for low-resource languages. M3T-Bench thus establishes a new standard for multilingual table understanding, offering both a challenging evaluation platform and a scalable methodology for future research.\footnote{https://github.com/sdxvv/m3TQA/tree/master}
\end{abstract}

\section{Introduction}
\label{sec:introduction}


Tabular data serves as a cornerstone of real-world information systems, underpinning critical applications from recommender systems~\citep{guo2017deepfm} to financial analytics~\citep{clements2020sequential}. The advent of large language models (LLMs) has catalyzed breakthroughs in table reasoning tasks, including table question answering (QA)~\citep{ye2023large,nahid2024tabsqlify,sui2024tap4llm,zhang2025alter,zhang2024reactable}, fact verification~\citep{ye2023large,zhang2024e5,nahid2024normtab}, and semantic retrieval~\citep{pourreza2023din,talaei2024chess}.

However, a critical linguistic bias persists: Current research predominantly targets English-language tables, neglecting the complexities of multilingual table understanding. While recent efforts have begun addressing this gap—such as MULTITAT~\citep{zhang2025multitat} for multilingual table-text QA, XINFOTABS~\citep{minhas2022xinfotabs} extending INFOTABS~\citep{gupta2020infotabs} to 10 languages, and TATA~\citep{gehrmann2023tata} for African languages—these initiatives remain limited. Existing datasets exhibit two fundamental limitations: (1) \textbf{Geolinguistic imbalance.}Existing datasets focus predominantly on Indo-European languages (Europe/North America/Middle East/Indian subcontinent) and high-impact languages (e.g., Chinese, Arabic), with inadequate coverage of most language families. (2) \textbf{Scale constraints.}Current cross-lingual datasets lack sufficient scale for granular linguistic feature analysis.

\begin{figure}[t!]
\begin{center}
    \includegraphics[width=0.45\textwidth]{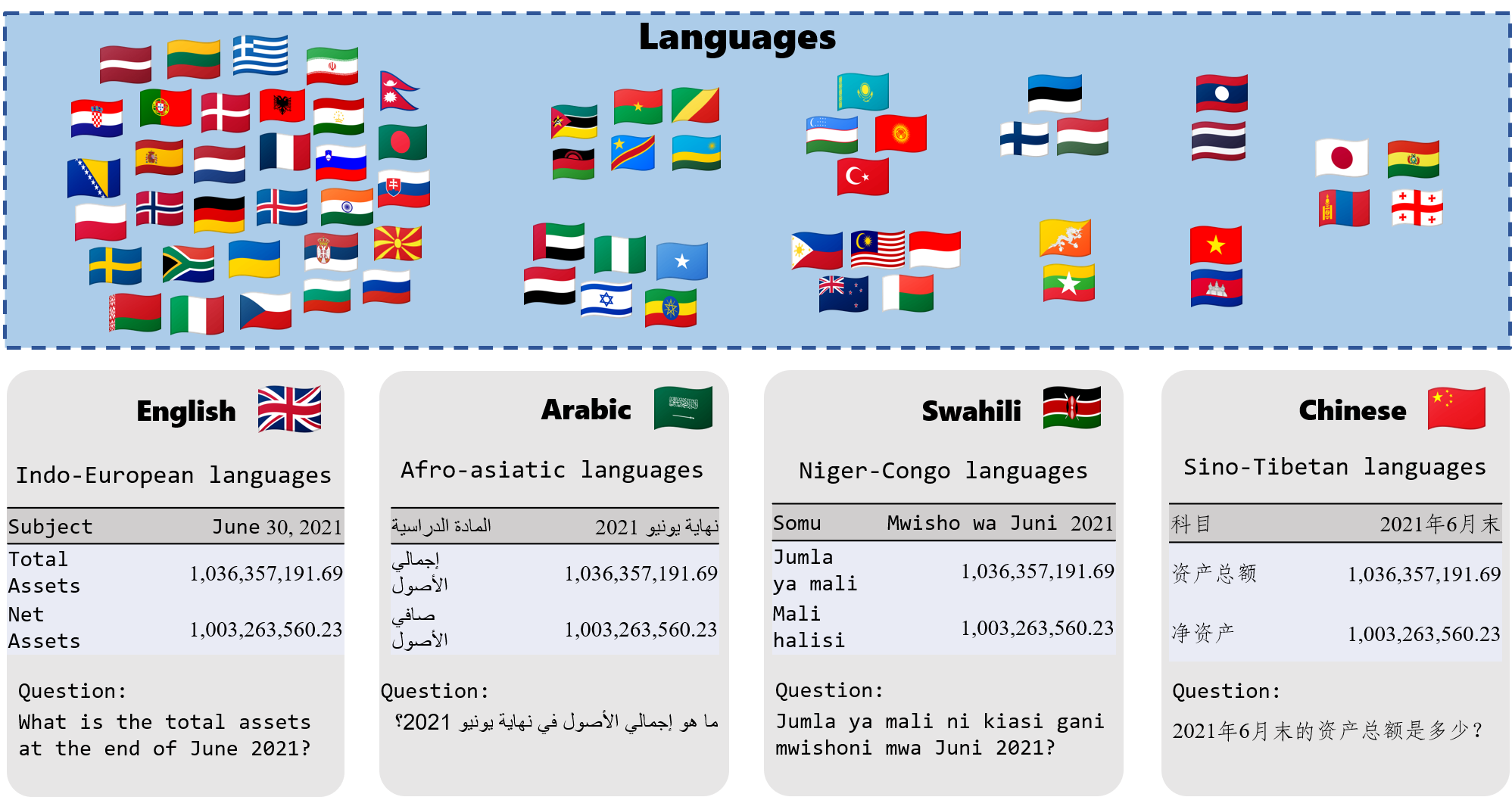}
    \caption{Partial language visualization from \benchmark{}. Spatially proximate languages belong to the same language family. Widely-used language families and their representative languages are shown below (Korean, Japanese, and other language isolates are categorized separately).}
    \label{fig:code_classes}
    \vspace{-15pt}
\end{center}
\end{figure}

To address these gaps, we present the Massive Multilingual Multitask Table Benchmark (M3T-Bench). In our work, we collected 50 real-world tables from various sources and languages (41 Chinese and 9 English) to ensure experimental generalizability. Adopting the four types of table structure from IM-TQAS~\citep{zheng2023tqa} as selection criteria, we incorporated rich-text tables to enhance complexity. We established a comprehensive translation and verification pipeline to extend the dataset to 97 languages, utilizing DeepSeek and GPT-4o as translation engines through a six-step process. Translation accuracy was rigorously validated via back-translation, with a median BLEU score of 60.19 (minimum: 34.06) achieved after excluding low quality translations. Finally, 2,916 professionally annotated QA pairs were generated to complete M3T-Bench. We conducted a series of experiments on it to evaluate diverse LLMs' tabular comprehension and cross-lingual capabilities, revealing that: {experimental conclusions}.

Our contributions are summarized as follows:
\begin{itemize}
    \item We introduce a large-scale multilingual table understanding dataset covering 97 languages. This significantly extends linguistic coverage beyond existing benchmarks.
    \item We propose an efficient translation pipeline using advanced LLMs. This pipeline achieves high-quality cross-lingual table conversion (median BLEU: 60.19).
    \item We designed four tasks on M3T-Bench to systematically evaluate various models’ capabilities in interpreting and processing tabular data. Our experiments demonstrate that synthetically generated, unannotated QA pairs enhance cross-lingual comprehension. Crucially, this finding reveals a viable pathway toward optimizing performance for low-resource languages.
\end{itemize}

\section{\benchmark{}}

\begin{figure*}[ht!]
\begin{center}
    \includegraphics[width=0.85\textwidth]{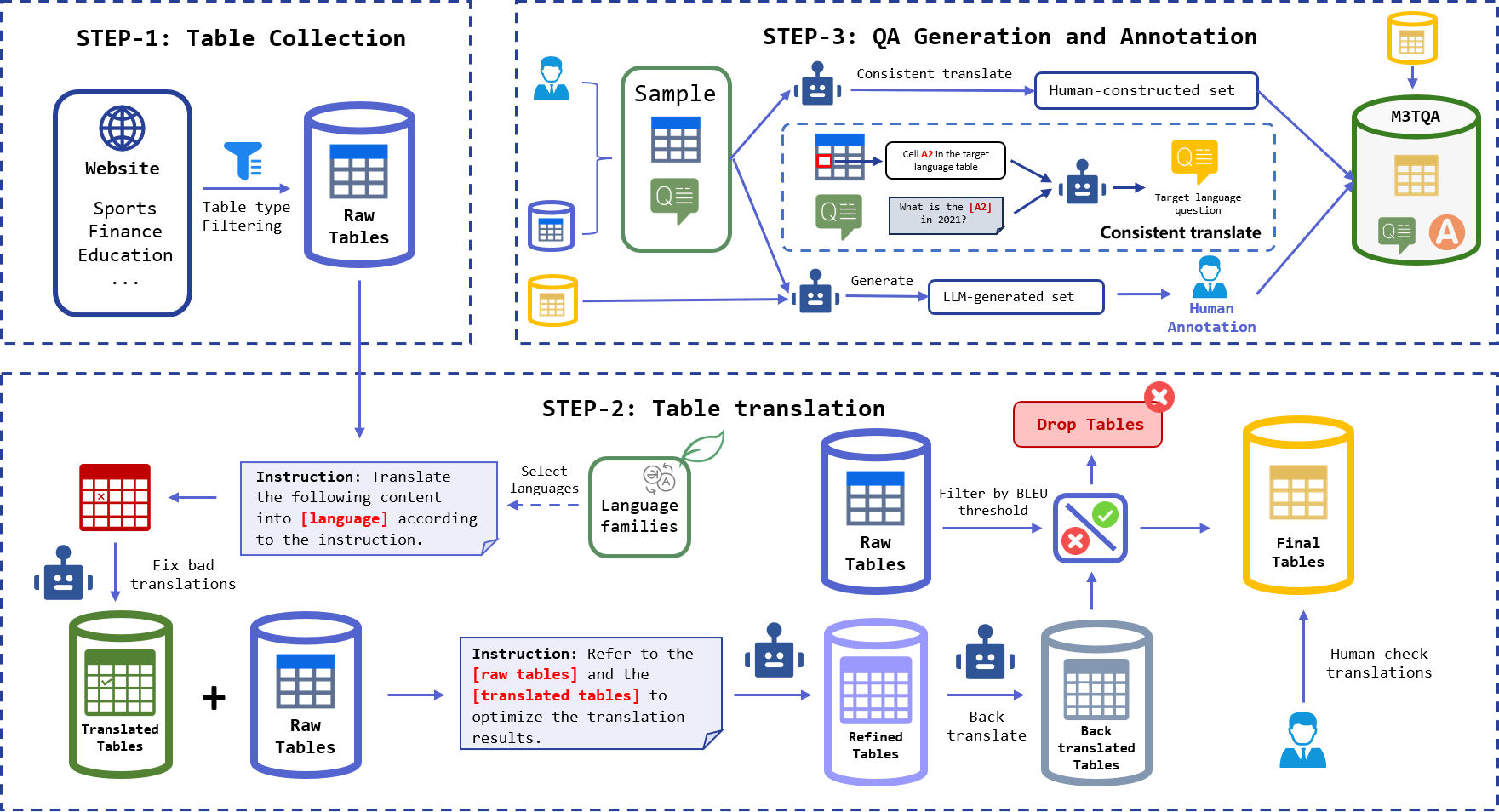}
    \caption{The framework of constructing \benchmark{}.}
    \label{fig:framework}
    \vspace{-15pt}
\end{center}
\end{figure*}

We present \benchmark{} (Massively Multilingual Multitask Table Question Answering), a curated dataset designed for multilingual multitask table QA. The dataset originates from 50 source tables in English and Chinese. Through a six-step procedure, we extend it to 97 languages. Question-answer pairs are constructed via a hybrid approach combining human annotation and LLM generation, as shown in Figure \ref{fig:framework}. This yields a linguistically diverse resource that authentically reflects real-world multilingual challenges in table QA tasks.
\subsection{Data Collection}
We collected 50 tables from real-world sources to ensure data authenticity and complexity. Our sources included public annual reports from China's Shanghai Stock Exchange and Shenzhen Stock Exchange, and statistical reports from Statistics Canada and the US National Science Foundation. Table selection followed the structural taxonomy established in prior work~\citep{wang2021tuta, zheng2023tqa}, categorizing tables into four types:
(1) \textbf{Relational tables} Table data have a vertical layout where the top rows contain column headers, and the rest hold the main content.
(2) \textbf{Entity tables} Table arrange data horizontally, with the first few columns showing row headers and the remaining cells forming the body.
(3) \textbf{Matrix tables} Table data shows relationships in both directions, usually with multi-level headers.
(4) \textbf{Composite tables} Table data contains headers in variable positions (beyond top/left edges), reflecting complex real-world layouts.

\begin{table}[t]
\centering
\resizebox{0.950\columnwidth}{!}{
\begin{tabular}{lccr}
    \toprule
    \multicolumn{4}{c}{\textbf{Table Size}} \\ \midrule
     & Max & Min & Mean \\
    \textbf{Rows} & 57 & 5 & 15.52 \\
    \textbf{Columns} & 15 & 5 & 8.30 \\ \midrule
    \multicolumn{4}{c}{\textbf{Table Type}} \\ \midrule
     &  & Number & Proportion \\
    \textbf{Relational Tables} &  & 25 & 50\% \\
    \textbf{Entity Tables} &  & 6 & 12\% \\
    \textbf{Matrix Tables} &  & 13 & 26\% \\
    \textbf{Composite Tables} &  & 6 & 12\% \\ \midrule
    \multicolumn{4}{c}{\textbf{Table Features}} \\ \midrule
    \textbf{Ratio of Numerical Cells} &  &  & 56.66\% \\
    \textbf{English Tables} &  &  & 9 \\
    \textbf{Chinese Tables} &  &  & 41 \\
    \textbf{Multilingual Tables} &  &  & 4,349 \\
    \bottomrule
\end{tabular}}
\caption{Data statistics of Tables. Relational tables have a vertical layout while Entity tables arrange data horizontally. Matrix tables show relationships in both directions. Composite tables contains headers in variable positions.}
\label{table1}
\end{table}

\subsection{Table translation}
To manage translation costs, we leveraged multilingual LLMs (DeepSeek and GPT-4o) for table translation through a six-phase pipeline:

(1) \textbf{Initial Translation}:
    Tabular data possesses inherent contextual features. To preserve structural semantics during translation, we serialize tables as two-dimensional lists. We then construct prompts to indicate LLM to translate content while strictly maintaining this matrix structure. Post-translation, the serialized output is reconstructed into the target-language table format.
    
(2) \textbf{Cell Correction}:
    At this stage, LLM is used to reprocess cells with translation defects to correct errors or translation omissions and ensure the integrity of the table in the output of the target language.
    
(3) \textbf{Global Refinement}:
    Let LLM analyze the original table and the translated table to determine whether the translation result can be refined and generate the refined result. The purpose of this stage is to correct the deviations that occurred during the translation process based on the overall semantics of the table.
    
(4) \textbf{Cell Refinement}:
    Similar to the previous stage, this stage refines the original content and translation results of each cell separately.
    
(5) \textbf{Back-Translation Validation}:
    The translated table is backtranslated into the original language using LLM and compared with the meaning of the original table. The comparison scheme uses an adaptive BLEU score for non-numeric cells:
    \[
    \text{BLEU}_{\text{cell}} = 
    \begin{cases} 
        \text{BLEU-1} & \text{if $L$} \leq 3 \\[3pt]
        \frac{\text{BLEU-1} + \text{BLEU-2}}{2} & \begin{array}{@{}l@{}} 
            \text{if } 4 \leq \text{$L$} \leq 7 
        \end{array} \\[8pt]
        \frac{1}{4}\sum_{n=1}^{4} \text{BLEU-}n & \text{if $L$} \geq 8 
    \end{cases}
    \]
    
    Here $L$ represents the text length of an individual table cell. The proposed adaptive BLEU computation scheme effectively mitigates the scoring biases inherent in conventional BLEU-2 and BLEU-4 metrics when evaluating short text sequences and reduces the undue penalization of brevity. The table score is the average of all non-numeric cell scores, and tables with scores below the threshold will be discarded.

    After filtering, approximately 10\% of tables with low BLEU scores were excluded. The final dataset contains 4,349 tables, with a median BLEU score of 60.19 and a minimum BLEU score of 34.06.
    
(6) \textbf{Human Check}:
    Check the remaining tables and correct any errors found.

Table\ref{table1} presents the statistical characteristics of the collected tabular data.


\begin{table}
\centering
\resizebox{.95\columnwidth}{!}{
\begin{tabular}{lccc}
    \toprule
    \textbf{Properties} & \makecell{\textbf{LLM-generated} \\ \textbf{test set}}  & \makecell{\textbf{Human-constructed} \\ \textbf{test set}} & \textbf{Entire test set} \\ \midrule
    \multicolumn{4}{c}{\textbf{\instruct{}}} \\ \midrule
    Numerical Computation & 28,592 &  & 28,592 \\
    Cell Extraction & 1,103 &  & 1,103 \\ 
    Factual Verification & 3,157 &  & 3,157 \\
    Open-Ended Questions & 7,171 &  & 7,171 \\
    Total & 39,077 &  & 39,077  \\ 
    \midrule
    \multicolumn{4}{c}{\textbf{\benchmark{}}} \\ \midrule
    
    Numerical Computation & 2,258 & 1,669 & 3,927 \\
    Cell Extraction & 337 & 1,254 & 1,591 \\ 
    Factual Verification & 48 & 289 & 337 \\
    Open-Ended Questions & 273 & 1,082 & 1,355 \\
    Total & 2,916 & 4,294 & 7,210 \\ \midrule
    Max Prompt Length & 11,248 & 10,932 & 11,248 \\
    Mean prompt Length & 3,888.03 & 3,768.97 & 3,817.12 \\ 
    \bottomrule
\end{tabular}}
\caption{Data statistics of QA sets}
\label{table2}
\end{table}

\subsection{Question-Answer Generation and Annotation}
\paragraph{Hybrid QA Generation}
Question-answer pairs were generated through a combined human-LLM approach to balance cost efficiency with data quality. Human annotators first created a seed corpus by manually generating one QA pair per table according to TableBench's\cite{wu2025tablebench} question classification schema. These seed examples were then incorporated into prompts to guide LLMs in expanding the dataset. It should be noted that an approach involving first constructing Question-Answer (QA) pairs and subsequently translating both the source table and the generated QA pairs into the target language may introduce inconsistencies. Specifically, identical content or entities within the table and QA pairs risk being translated differently due to factors such as lack of shared context during the separate translation stages. To resolve this inconsistency, the QA pairs are instead generated directly from the source-language tables. This hybrid methodology yielded an initial dataset of 10 QA pairs per table-language combination.

\paragraph{Data Partitioning and Annotation}
For test set construction, we randomly selected one QA pair per table-language combination. Annotators screened, inspected, and refined these pairs to meet task requirements, excluding lower-quality pairs. This process resulted in a curated test set of 2,916 QA pairs, averaging approximately 30 samples per language. The remaining LLM-generated QA pairs formed the training set \instruct{}, totaling 39,077 pairs. 

Concurrently, we extended the human-constructed English QA set to all target languages. To maintain table consistency during translation, we annotated the corresponding entities in the table when constructing question-answer pairs, and extracted the entities in the question and answer before translation, associating them with the corresponding entries in the target language table. This entity alignment preserved semantic equivalence across language transformations. This yielded a curated test set of 4,294 QA pairs.

Table \ref{table2} presents relevant information regarding the constructed Question-Answer (QA) pairs.

\paragraph{Quality Control}
To ensure m$^3$T-Bench's reliability, we implemented rigorous quality controls throughout the 50-day annotation cycle. All annotators possessed graduate-level expertise and full Chinese/English fluency to comprehend source tables accurately. Quality control strategies included comprehensive training with open question channels, two interim quality checks with real-time issue resolution, and final random sampling verification. This multi-stage oversight guaranteed consistent adherence to annotation standards while addressing challenges inherent in multilingual table comprehension.

\subsection{Task Definition}
\label{sec:Task Definition}
Our task is defined as:
Given the k-th language $L_k\in\{L_i\}^K_{i=1}$, where $K=97$, indicating the number of languages. For a table QA task, we provide a table $T^{L_k}$ of the language $L_k$ containing $R$ rows and $C$ columns, as well as the description of the question $q^{L_k}$, as input of LLM $M$, so that $M$ generates the corresponding result $a^{L_k}$. The process can be expressed as $a=M(T,q)$. Then use the evaluation function $I(\cdot)$ to evaluate the answer of $M$, that is, $s^{L_k}=I(a^{L_k},y^{L_k})$. In summary, the process can be described as:
$$s^{L_k}=I(M(T^{L_k},q^{L_k}),y^{L_k})$$

Specifically, tabular question answering tasks can be categorized into four types based on answer characteristics:

(1) \textbf{Numerical Computation}: This category involves numerical operations such as aggregation, counting, or calculations based on tabular data. To address potential variations in numerical representations across languages, responses $a^{L_k}$must strictly adhere to Arabic numerals. The assessment metric computes the Jaccard similarity coefficient between the model-generated answer ($A_{\text{gen}}$) and ground truth answer ($A_{\text{gt}}$), defined as:
    \[
    \mathcal{J}(A_{\text{gen}}, A_{\text{gt}}) = \frac{|A_{\text{gen}} \cap A_{\text{gt}}|}{|A_{\text{gen}} \cup A_{\text{gt}}|}
    \]
    
(2) \textbf{Cell Extraction}: Answers require retrieving content from one or multiple table cells. To mitigate language differences in interpretation, the model also needs to answer the coordinates of the cell when returning the cell content. The evaluation function computes the Jaccard similarity coefficient between the model-generated answer and the ground truth.

(3) \textbf{Factual Verification}: Answers are restricted to binary outcomes (yes/no). Despite language-specific expressions for affirmation/negation, models are required to output standardized responses ``T'' or ``F''. The evaluation employs the F1-score to assess prediction accuracy against ground truth labels.

(4) \textbf{Open-Ended Questions}: For open-ended questions requiring descriptive answers, we employ the ROUGE-L metric to assess the similarity between the model-generated answer $a^{L_k}$ and the ground truth $y^{L_k}$.

The overall system performance $S$ aggregates all QA scores:

\[
S = \frac{1}{|\mathcal{D}_{\text{test}}|} \sum_{(q,a) \in \mathcal{D}_{\text{test}}} s^{(q,a)}
\]

where $\mathcal{D}_{\text{test}}$ is the test set, and $s^{(q,a)}$ is the score for an individual question-answer pair in the test set.

\section{Method}
In recent years, numerous task-specific optimization strategies have been proposed for large language models, such as supervised fine-tuning (SFT) and reinforcement learning (RL), and demonstrated exceptional performance in various tasks. This section describes multiple learning and training methods for the \benchmark{} task that enable lower-parameter LLMs to generalize across multiple languages.
\subsection{Thinking}
Early LLMs relied on rapid response mechanisms and intuitive judgments to achieve efficient but shallow task processing. However, this approach becomes inadequate when confronted with increasingly complex task requirements. With the emergence of thinking paradigms like Chain-of-Thought (CoT), an increasing number of LLMs now support thinking modes. Nevertheless, when encountering unseen languages, LLMs frequently exhibit thinking disorganization or infinite loops, which impairs their problem-solving capabilities. To address this, we integrate the model's thinking process into training, utilizing the thinking paradigms of large-parameter multilingual LLMs to optimize the thinking logic of smaller-parameter LLMs.

\begin{figure}[ht!]
\begin{center}
    \includegraphics[width=0.4\textwidth]{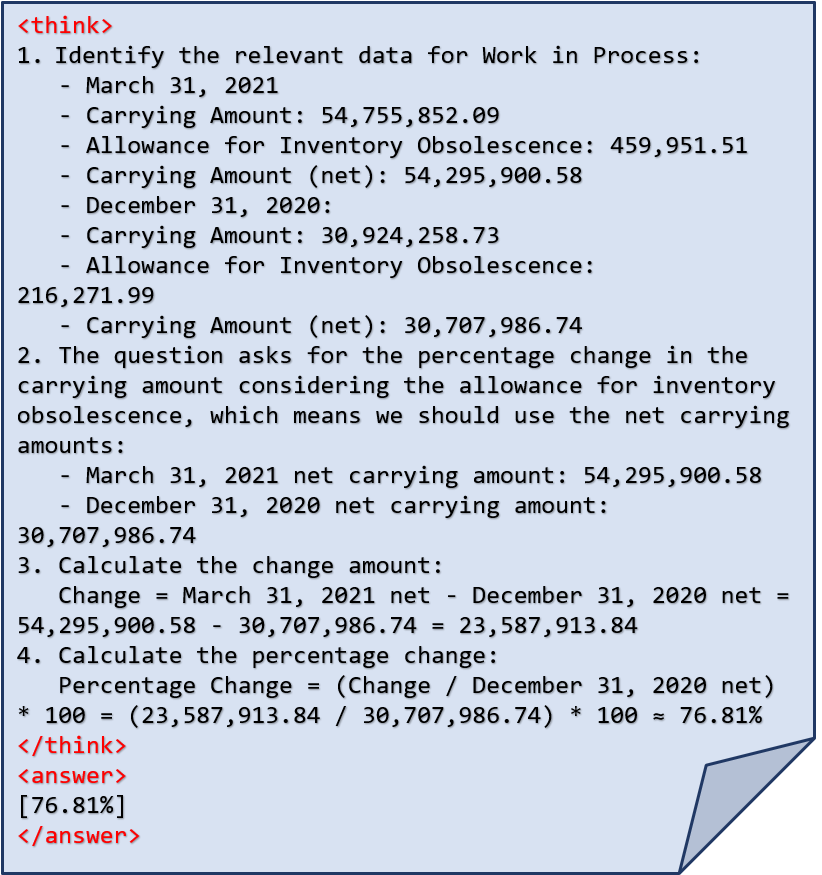}
    \caption{An example of our thinking training paradigm. Model thinking traces are captured within \texttt{<think>} tags and concatenated with final predictions in \texttt{<answer>} tags.}
    \label{fig:think-format}
    \vspace{-15pt}
\end{center}
\end{figure}

Specifically, we utilize task-specific prompts to direct both DeepSeek-R1 and GPT-4o to output their thinking traces within dedicated \texttt{<think>} tags. These traces are then concatenated with corresponding model-generated outputs enclosed within <answer> tags. This composite structure forms the training objective for subsequent model fine-tuning, as illustrated in Figure \ref{fig:think-format}.

\subsection{Supervised Fine-Tuning}
We perform instruction fine-tuning on \instruct{} for table question answering, minimizing $\mathcal{L}$:
\[
\mathcal{L} = \mathbb{E}_{(q^{L_k}, a^{L_k}) \sim \mathcal{D}} \left[ -\log P(a^{L_k} \mid T^{L_k}, q^{L_k}; \theta) \right]
\]
where $q^{L_k}$ and $a^{L_k}$ denote the question and answer in language $L_k$, $K = 97$ represents the total number of languages, and $\mathcal{D}$ is the training data distribution

Experiments are conducted for both thinking and non-thinking modes:

\textbf{Non-Thinking Mode}: $a^{L_k}$ consists of the answer label \texttt{<answer>} followed by the ground truth answer.

\textbf{Thinking Mode}: $a^{L_k}$ contains the thinking label \texttt{<think>}, followed by the thinking process, and concludes with \texttt{<answer>} and the ground truth answer.

\subsection{Group Relative Policy Optimization}
Building upon the optimal SFT checkpoint, we apply GRPO~\citep{shao2024deepseekmath} optimization for reinforcement learning. The scoring mechanism for both thinking and non-thinking modes consistently employs the evaluation function $I(\cdot)$ established in task definition.

\section{Experiments}
To ensure fair evaluation across diverse tasks, we designed unified prompt templates for each task category. Model outputs were subjected to strict normalization procedures to extract final answers while eliminating confounding factors from extraneous information in LLM responses. All experiments were conducted on the \benchmark{}. We fine-tuned multiple open-source LLMs using \instruct{}, with further optimization via GRPO.

\begin{table*}[!ht]
\centering
\resizebox{0.92 \textwidth}{!}{
\begin{tabular}{l|cccccccccccccccccccc}
\toprule
\rowcolor{red!10} & \multicolumn{5}{c}{\textbf{Afroasiatic languages}} & \multicolumn{5}{c}{\textbf{Indo-European languages}} & \multicolumn{5}{c}{\textbf{Austronesian languages}}  & \multicolumn{5}{c}{\textbf{Sino-Tibetan languages}} \\ 
\rowcolor{red!10} & NC & CE & FV & OEQ & Avg. & NC & CE & FV & OEQ & Avg. & NC & CE & FV & OEQ & Avg. & NC & CE & FV & OEQ & Avg.  \\ \midrule
\rowcolor{yellow!20} \multicolumn{21}{c}{\textbf{Open-source Methods}} \\ \midrule
   \rowcolor{green!10} Baichuan2-7B-Chat & 0.00 & 0.38 & 3.57 & 0.93 & 0.42 & 0.10 & 0.47 & 3.77 & 2.67 & 0.82 & 0.00 & 1.11 & 3.85 & 1.72 & 0.78 & 1.18 & 0.00 & 0.00 & 1.37 & 0.86 \\ 
   \rowcolor{green!10} Deepseek-llm-7B-Chat & 0.34 & 0.02 & 3.57 & 3.27 & 0.95 & 0.67 & 0.24 & 8.18 & 2.93 & 1.32 & 0.86 & 0.25 & 7.69 & 3.95 & 1.67 & 2.16 & 0.00 & 0.00 & 0.00 & 1.17 \\
   \rowcolor{green!10} Seed-Coder-8B-Instruct & 4.01 & 2.09 & 53.57 & 18.75 & 8.60 & 9.21 & 4.26 & 61.64 & 23.31 & 12.99 & 6.82 & 3.94 & 53.85 & 25.63 & 12.21 & 11.96 & 3.42 & 62.50 & 6.41 & 11.53 \\ 
   \rowcolor{green!10} Llama-3-8B-Instruct & 4.26 & 1.50 & 7.14 & 24.10 & 7.32 & 6.11 & 1.18 & 17.61 & 29.89 & 9.96 & 4.42 & 1.03 & 23.08 & 33.39 & 10.15 & 6.27 & 1.28 & 12.50 & 6.41 & 5.37 \\
   \rowcolor{green!10} Llama-3.1-8B-Instruct & 1.53 & 1.99 & 10.71 & 22.06 & 5.75 & 3.60 & 1.92 & 13.21 & 28.84 & 8.35 & 1.52 & 1.47 & 23.08 & 30.24 & 8.11 & 1.76 & 0.00 & 25.00 & 10.26 & 3.86 \\
   \rowcolor{green!10} Qwen3-8B-nothinking & 6.57 & 2.56 & 78.57 & 26.69 & 12.72 & 6.66 & 4.03 & 68.55 & 30.80 & 13.22 & 5.52 & 3.92 & 65.38 & 36.07 & 14.11 & 2.35 & 3.23 & 75.00 & 5.27 & 6.74 \\
   \rowcolor{green!10} Qwen3-8B-thinking & 27.08 & 7.89 & 50.00 & 19.30 & 22.40 & 39.45 & 10.32 & 52.83 & 24.39 & 30.84 & 32.89 & 6.80 & 61.54 & 28.91 & 27.71 & 35.10 & 5.56 & 62.50 & 4.62 & 24.30 \\
   \rowcolor{green!10} GLM-4-9B-0414-nothinking & 9.88 & 1.63 & 3.57 & 21.36 & 9.74 & 16.68 & 2.48 & 7.55 & 25.07 & 14.76 & 9.06 & 2.62 & 19.23 & 29.75 & 12.08 & 13.79 & 1.52 & 12.50 & 3.13 & 8.98 \\
   \rowcolor{green!10} GLM-4-9B-0414-thinking & 17.21 & 1.13 & 35.71 & 17.96 & 14.55 & 26.18 & 2.30 & 47.17 & 20.68 & 20.83 & 22.65 & 3.10 & 38.46 & 22.71 & 19.05 & 20.12 & 2.56 & 37.50 & 4.17 & 14.11 \\
   \rowcolor{green!10} Phi-4-14B & 21.81 & 1.50 & 57.14 & 23.10 & 19.09 & 37.26 & 2.18 & 52.83 & 28.14 & 28.55 & 32.46 & 2.76 & 46.15 & 32.33 & 26.41 & 26.65 & 1.50 & 50.00 & 4.38 & 18.05 \\
   \rowcolor{green!10} Moonlight-16B-A3B-Instruct & 2.37 & 0.59 & 21.43 & 8.04 & 3.90 & 3.54 & 2.93 & 18.87 & 6.46 & 4.61 & 3.18 & 2.00 & 30.77 & 3.60 & 4.43 & 3.33 & 0.51 & 25.00 & 2.16 & 3.55 \\
   \rowcolor{green!10} ERNIE-4.5-21B-A3B-PT & 16.24 & 5.89 & 75.00 & 26.47 & 18.53 & 25.73 & 4.57 & 66.67 & 29.27 & 23.51 & 24.82 & 6.76 & 61.54 & 35.47 & 24.67 & 21.76 & 2.56 & 75.00 & 7.38 & 17.42 \\
   \rowcolor{green!10} GLM-4-32B & 31.71 & 10.13 & 64.29 & 19.68 & 26.19 & 42.08 & 10.31 & 65.41 & 23.99 & 32.76 & 41.54 & 9.54 & 69.23 & 27.41 & 33.03 & 37.65 & 6.15 & 62.50 & 5.17 & 25.92 \\
   \rowcolor{green!10} Qwen3-32B-nothinking & 18.00 & 4.80 & 67.86 & 21.21 & 17.96 & 20.22 & 8.23 & 79.25 & 25.84 & 21.16 & 11.01 & 5.81 & 69.23 & 30.42 & 16.57 & 12.65 & 10.21 & 87.50 & 5.43 & 14.71 \\
   \rowcolor{green!10} Qwen3-32B-thinking & 37.23 & 27.27 & 67.86 & 20.27 & 33.39 & 53.67 & 29.58 & 73.58 & 22.35 & 43.41 & 47.25 & 23.92 & 76.92 & 26.81 & 39.61 & 46.01 & 25.79 & 75.00 & 5.78 & 36.06 \\ \midrule
\rowcolor{yellow!20} \multicolumn{21}{c}{\textbf{Close-source Methods}} \\ \midrule
   \rowcolor{blue!10} GPT-3.5-turbo & 9.00 & 8.14 & 67.86 & 27.66 & 14.96 & 11.45 & 7.96 & 78.62 & 33.11 & 17.57 & 8.26 & 10.09 & 65.38 & 39.84 & 17.66 & 7.65 & 5.50 & 87.50 & 7.73 & 11.14 \\
   \rowcolor{blue!10} GPT-4o & 37.94 & 20.63 & 46.43 & 24.07 & 31.84 & 43.32 & 25.92 & 48.15 & 28.84 & 36.97 & 44.01 & 20.07 & 42.31 & 33.92 & 36.58 & 28.82 & 18.90 & 50.00 & 6.84 & 23.80 \\
   \rowcolor{blue!10} Gemini-2.0 & 27.86 & 28.04 & 71.43 & 24.59 & 29.42 & 31.95 & 27.37 & 59.12 & 17.91 & 29.49 & 37.24 & 21.78 & 65.38 & 19.44 & 31.81 & 27.84 & 30.00 & 87.50 & 5.52 & 27.86 \\
   \rowcolor{blue!10} Gemini-2.5 & 50.57 & 39.73 & 72.41 & 19.20 & 43.29 & 60.42 & 53.20 & 76.58 & 27.06 & 53.30 & 61.75 & 46.73 & 73.08 & 31.61 & 53.18 & 51.60 & 44.44 & 87.50 & 5.91 & 44.37 \\
   \rowcolor{blue!10} Deepseek-Chat-V3 & 46.13 & 15.60 & 67.86 & 28.33 & 36.89 & 56.29 & 20.66 & 66.04 & 31.91 & 44.32 & 57.92 & 21.12 & 50.00 & 38.63 & 45.50 & 45.49 & 16.82 & 87.50 & 9.24 & 34.46 \\ \midrule
\rowcolor{yellow!20} \multicolumn{21}{c}{\textbf{Open-Source Training Methods}} \\ \midrule
   \rowcolor{pink!10} Llama-3.1-8B-Instruct-SFT & 30.32 & 18.21 & 85.71 & 25.66 & 29.38 & 44.68 & 19.03 & 77.99 & 27.67 & 37.31 & 39.61 & 14.92 & 76.92 & 33.30 & 34.75 & 26.56 & 14.40 & 100.00 & 6.28 & 24.05 \\
   \rowcolor{pink!10} Llama-3.1-8B-Instruct-SFT-GRPO & 51.39 & 44.67 & 66.67 & 33.94 & 47.97 & 53.63 & 48.34 & 51.87 & 43.09 & 50.43 & 40.60 & 51.30 & 43.21 & 45.54 & 44.18 & 50.49 & 44.29 & 34.60 & 44.65 & 47.14 \\
   \rowcolor{pink!10} Qwen3-8B-SFT & 41.56 & 17.46 & 53.57 & 25.16 & 33.69 & 49.49 & 22.59 & 62.26 & 29.42 & 40.41 & 47.31 & 20.75 & 73.08 & 34.05 & 40.10 & 42.16 & 9.98 & 75.00 & 7.08 & 30.25 \\
   \rowcolor{pink!10} Qwen3-8B-SFT-GRPO & 54.73 & 33.50 & 44.44 & 30.24 & 46.41 & 53.36 & 34.75 & 54.72 & 36.86 & 46.27 & 45.49 & 20.88 & 62.50 & 8.02 & 34.28 & 54.43 & 36.31 & 65.38 & 41.87 & 48.49 \\

\midrule
\rowcolor{red!10} & \multicolumn{5}{c}{\textbf{Austroasiatic languages}} & \multicolumn{5}{c}{\textbf{Niger-Congo languages}} & \multicolumn{5}{c}{\textbf{Uralic languages}}  & \multicolumn{5}{c}{\textbf{Mongolic languages}} \\ 
\rowcolor{red!10} & NC & CE & FV & OEQ & Avg. & NC & CE & FV & OEQ & Avg. & NC & CE & FV & OEQ & Avg. & NC & CE & FV & OEQ & Avg.  \\ \midrule
\rowcolor{yellow!20} \multicolumn{21}{c}{\textbf{Open-source Methods}} \\ \midrule
   \rowcolor{green!10} Baichuan2-7B-Chat & 0.00 & 0.00 & 0.00 & 0.47 & 0.08 & 0.38 & 0.23 & 0.00 & 0.81 & 0.41 & 0.00 & 0.20 & 0.00 & 1.00 & 0.19 & 0.00 & 0.00 & 0.00 & 0.94 & 0.18 \\
   \rowcolor{green!10} Deepseek-llm-7B-Chat & 0.00 & 0.00 & 0.00 & 0.00 & 0.00 & 0.00 & 0.00 & 15.00 & 3.13 & 1.73 & 0.01 & 0.00 & 0.00 & 2.05 & 0.32 & 0.00 & 0.00 & 0.00 & 1.42 & 0.27 \\ 
   \rowcolor{green!10} Seed-Coder-8B-Instruct & 3.81 & 1.33 & 50.00 & 2.02 & 5.78 & 1.77 & 2.47 & 45.00 & 22.27 & 9.46 & 8.98 & 3.09 & 55.56 & 14.92 & 10.41 & 3.01 & 3.42 & 57.14 & 15.30 & 8.02 \\ 
   \rowcolor{green!10} Llama-3-8B-Instruct & 0.95 & 2.22 & 0.00 & 1.82 & 1.33 & 3.79 & 1.16 & 20.00 & 25.27 & 8.97 & 1.06 & 1.09 & 11.11 & 20.12 & 4.31 & 3.41 & 1.85 & 14.29 & 17.82 & 6.35 \\
   \rowcolor{green!10} Llama-3.1-8B-Instruct & 2.86 & 0.00 & 0.00 & 4.62 & 2.32 & 2.84 & 2.18 & 25.00 & 24.79 & 9.04 & 1.33 & 1.43 & 11.11 & 22.07 & 4.84 & 1.61 & 9.38 & 28.57 & 26.46 & 9.04 \\
   \rowcolor{green!10} Qwen3-8B-nothinking & 7.62 & 1.33 & 75.00 & 16.33 & 11.79 & 2.02 & 1.53 & 70.00 & 27.72 & 12.28 & 5.00 & 1.94 & 55.56 & 25.16 & 9.29 & 9.64 & 7.41 & 28.57 & 23.42 & 12.72 \\
   \rowcolor{green!10} Qwen3-8B-thinking & 28.56 & 11.76 & 75.00 & 19.87 & 26.15 & 16.46 & 8.70 & 29.63 & 16.12 & 15.47 & 43.82 & 5.51 & 55.56 & 21.64 & 33.24 & 36.75 & 14.81 & 57.14 & 21.53 & 30.77 \\
   \rowcolor{green!10} GLM-4-9B-nothinking & 3.33 & 4.17 & 0.00 & 6.49 & 3.85 & 3.79 & 2.57 & 5.00 & 21.39 & 7.43 & 15.44 & 1.90 & 11.11 & 20.62 & 13.36 & 8.84 & 1.17 & 14.29 & 21.43 & 10.02 \\
   \rowcolor{green!10} GLM-4-9B-thinking & 11.90 & 0.00 & 25.00 & 3.40 & 8.52 & 7.05 & 3.47 & 30.00 & 15.63 & 9.64 & 30.24 & 3.23 & 44.44 & 15.95 & 23.21 & 20.88 & 0.00 & 28.57 & 17.87 & 16.78 \\
   \rowcolor{green!10} Phi-4-14B & 27.89 & 1.61 & 57.14 & 28.28 & 23.82 & 15.23 & 2.45 & 55.56 & 20.89 & 16.11 & 36.69 & 3.06 & 44.44 & 27.35 & 28.84 & 25.30 & 0.00 & 42.86 & 17.10 & 19.87 \\
   \rowcolor{green!10} Moonlight-16B-A3B-Instruct & 0.06 & 2.15 & 14.29 & 1.23 & 1.40 & 0.74 & 1.67 & 25.00 & 6.22 & 3.97 & 4.03 & 1.56 & 33.33 & 2.21 & 4.34 & 0.30 & 1.98 & 42.86 & 7.45 & 4.03 \\
   \rowcolor{green!10} ERNIE-4.5-21B-A3B-PT & 17.59 & 1.61 & 85.71 & 25.02 & 18.88 & 16.72 & 2.57 & 72.22 & 22.76 & 18.46 & 24.24 & 1.02 & 66.67 & 23.82 & 21.09 & 19.68 & 3.91 & 42.86 & 24.50 & 18.75 \\
   \rowcolor{green!10} GLM-4-32B & 32.56 & 8.06 & 71.43 & 23.42 & 27.51 & 22.99 & 8.59 & 47.22 & 18.56 & 20.09 & 46.24 & 13.27 & 55.56 & 19.50 & 35.95 & 31.65 & 3.70 & 71.43 & 19.89 & 26.14 \\
   \rowcolor{green!10} Qwen3-32B-nothinking & 10.00 & 8.89 & 100.00 & 10.06 & 15.29 & 9.28 & 4.16 & 65.00 & 24.49 & 15.22 & 14.72 & 6.88 & 88.89 & 22.31 & 17.02 & 14.46 & 16.39 & 57.14 & 21.15 & 18.15 \\
   \rowcolor{green!10} Qwen3-32B-thinking & 34.29 & 42.06 & 75.00 & 12.23 & 34.85 & 25.38 & 14.27 & 40.00 & 22.61 & 23.00 & 56.38 & 26.68 & 77.78 & 18.17 & 45.45 & 47.43 & 33.64 & 71.43 & 20.66 & 40.99 \\ \midrule
\rowcolor{yellow!20} \multicolumn{21}{c}{\textbf{Close-source Methods}} \\ \midrule
   \rowcolor{blue!10} GPT-3.5-turbo & 9.35 & 3.33 & 85.71 & 32.62 & 16.15 & 6.04 & 8.41 & 61.11 & 25.04 & 14.79 & 11.13 & 8.08 & 88.89 & 29.68 & 16.12 & 9.33 & 5.36 & 85.71 & 24.16 & 14.94 \\
   \rowcolor{blue!10} GPT-4o & 40.68 & 25.81 & 42.86 & 27.77 & 35.19 & 36.16 & 21.20 & 44.44 & 25.79 & 30.59 & 39.25 & 24.78 & 44.44 & 30.26 & 35.06 & 46.51 & 22.92 & 42.86 & 25.03 & 37.71 \\
   \rowcolor{blue!10} Gemini-2.0 & 25.69 & 32.03 & 71.43 & 18.22 & 27.78 & 34.49 & 23.66 & 47.22 & 19.75 & 29.46 & 30.44 & 30.36 & 66.67 & 8.81 & 28.49 & 30.72 & 35.93 & 57.14 & 11.71 & 29.42 \\
   \rowcolor{blue!10} Gemini-2.5 & 51.46 & 54.33 & 100.00 & 21.48 & 48.67 & 46.86 & 47.40 & 55.56 & 17.38 & 41.40 & 62.01 & 55.85 & 88.89 & 30.05 & 56.94 & 51.61 & 47.53 & 71.43 & 25.23 & 46.86 \\
   \rowcolor{blue!10} Deepseek-Chat-V3 & 46.78 & 15.30 & 85.71 & 26.67 & 38.16 & 38.45 & 18.56 & 52.78 & 28.02 & 32.01 & 56.53 & 18.61 & 77.78 & 34.00 & 45.98 & 53.76 & 26.99 & 57.14 & 27.08 & 43.70 \\ \midrule

\rowcolor{yellow!20} \multicolumn{21}{c}{\textbf{Open-Source Training Methods}} \\ \midrule
   \rowcolor{pink!10} Llama-3.1-8B-Instruct-SFT & 17.71 & 10.56 & 75.00 & 9.24 & 18.15 & 26.20 & 9.61 & 65.00 & 28.45 & 25.24 & 41.07 & 15.71 & 66.67 & 26.57 & 34.75 & 36.75 & 18.89 & 71.43 & 22.56 & 32.43 \\
   \rowcolor{pink!10} Llama-3.1-8B-Instruct-SFT-GRPO & 47.09 & 38.39 & 60.31 & 41.47 & 44.62 & 55.88 & 50.79 & 58.29 & 36.85 & 51.04 & 56.81 & 46.65 & 55.24 & 38.87 & 51.03 & 45.99 & 43.52 & 58.39 & 44.39 & 45.71 \\
   \rowcolor{pink!10} Qwen3-8B-SFT & 40.12 & 23.04 & 85.71 & 29.14 & 36.60 & 30.45 & 18.12 & 66.67 & 24.51 & 28.66 & 48.87 & 28.42 & 55.56 & 24.53 & 41.35 & 48.59 & 24.54 & 71.43 & 26.07 & 40.97 \\
   \rowcolor{pink!10} Qwen3-8B-SFT-GRPO & 32.86 & 25.39 & 41.67 & 32.91 & 31.58 & 57.88 & 35.91 & 33.33 & 31.38 & 46.94 & 54.32 & 29.71 & 52.38 & 18.80 & 41.98 & 54.59 & 37.60 & 58.33 & 45.16 & 49.69 \\

\midrule
\rowcolor{red!10} & \multicolumn{5}{c}{\textbf{Dravidian languages}} & \multicolumn{5}{c}{\textbf{Turkic languages}} & \multicolumn{5}{c}{\textbf{Kra-Dai languages}}  & \multicolumn{5}{c}{\textbf{Language isolate}} \\ 
\rowcolor{red!10} & NC & CE & FV & OEQ & Avg. & NC & CE & FV & OEQ & Avg. & NC & CE & FV & OEQ & Avg. & NC & CE & FV & OEQ & Avg.  \\ \midrule
\rowcolor{yellow!20} \multicolumn{21}{c}{\textbf{Open-source Methods}} \\ \midrule
   \rowcolor{green!10} Baichuan2-7B-Chat &0.00 & 0.00 & 16.67 & 1.23 & 0.87 & 0.00 & 0.95 & 0.00 & 1.08 & 0.41 & 0.00 & 1.08 & 0.00 & 0.00 & 0.25 & 0.00 & 0.09 & 4.76 & 0.20 & 0.30 \\
   \rowcolor{green!10} Deepseek-llm-7B-Chat & 0.70 & 0.00 & 0.00 & 0.87 & 0.56 & 0.00 & 0.95 & 8.33 & 0.41 & 0.57 & 0.00 & 0.00 & 33.33 & 2.73 & 2.85 & 0.56 & 0.76 & 4.76 & 1.27 & 0.96 \\ 
   \rowcolor{green!10} Seed-Coder-8B-Instruct & 5.76 & 2.94 & 50.00 & 17.56 & 8.89 & 4.29 & 3.59 & 75.00 & 20.96 & 9.93 & 6.12 & 4.03 & 55.56 & 16.95 & 11.46 & 7.36 & 5.56 & 52.38 & 12.44 & 10.27 \\ 
   \rowcolor{green!10} Llama-3-8B-Instruct & 6.25 & 0.81 & 16.67 & 26.12 & 8.83 & 7.65 & 2.67 & 16.67 & 22.09 & 9.81 & 5.73 & 0.00 & 0.00 & 15.30 & 6.30 & 4.69 & 0.95 & 23.81 & 14.79 & 6.85 \\
   \rowcolor{green!10} Llama-3.1-8B-Instruct & 0.00 & 0.54 & 0.00 & 31.99 & 5.41 & 3.67 & 2.01 & 0.00 & 19.49 & 6.34 & 1.56 & 3.23 & 33.33 & 10.09 & 6.05 & 3.30 & 1.62 & 23.81 & 15.19 & 6.33 \\
   \rowcolor{green!10} Qwen3-8B-nothinking & 5.39 & 5.04 & 75.00 & 38.21 & 13.52 & 5.57 & 3.26 & 50.00 & 25.31 & 10.57 & 7.66 & 2.80 & 77.78 & 10.00 & 11.74 & 8.86 & 4.26 & 61.90 & 16.26 & 12.04 \\
   \rowcolor{green!10} Qwen3-8B-thinking & 31.92 & 9.46 & 50.00 & 30.89 & 27.86 & 39.25 & 9.95 & 58.33 & 17.75 & 29.67 & 33.07 & 6.99 & 44.44 & 15.63 & 23.78 & 35.14 & 14.56 & 47.62 & 14.04 & 27.23 \\
   \rowcolor{green!10} GLM-4-9B-0414-nothinking & 9.49 & 2.54 & 0.00 & 27.28 & 10.63 & 12.48 & 3.46 & 0.00 & 19.20 & 11.53 & 7.73 & 2.44 & 11.11 & 7.84 & 6.78 & 15.54 & 3.60 & 9.52 & 14.59 & 12.50 \\
   \rowcolor{green!10} GLM-4-9B-0414-thinking & 12.81 & 0.81 & 8.33 & 17.93 & 11.02 & 25.62 & 2.67 & 50.00 & 14.38 & 19.56 & 9.11 & 6.45 & 44.44 & 9.91 & 11.03 & 25.38 & 4.36 & 57.14 & 8.72 & 19.23 \\
   \rowcolor{green!10} Phi-4-14B & 38.07 & 0.86 & 50.00 & 30.00 & 29.57 & 35.32 & 4.27 & 40.00 & 22.81 & 26.59 & 26.61 & 1.94 & 66.67 & 11.68 & 20.12 & 34.39 & 2.94 & 61.90 & 15.04 & 25.27 \\
   \rowcolor{green!10} Moonlight-16B-A3B-Instruct & 0.33 & 0.97 & 25.00 & 3.61 & 1.98 & 1.69 & 1.70 & 20.00 & 4.34 & 2.99 & 0.34 & 1.12 & 11.11 & 0.91 & 1.37 & 2.94 & 2.94 & 33.33 & 6.03 & 5.10 \\
   \rowcolor{green!10} ERNIE-4.5-21B-A3B-PT & 26.92 & 5.02 & 58.33 & 30.85 & 24.32 & 24.83 & 5.75 & 60.00 & 24.54 & 22.31 & 29.87 & 3.41 & 88.89 & 12.37 & 23.63 & 22.73 & 3.30 & 57.14 & 14.37 & 18.69 \\
   \rowcolor{green!10} GLM-4-32B & 32.51 & 12.43 & 66.67 & 24.72 & 28.46 & 41.35 & 6.11 & 53.33 & 18.55 & 29.95 & 25.78 & 0.00 & 66.67 & 10.09 & 18.92 & 41.81 & 12.45 & 57.14 & 12.95 & 30.64 \\
   \rowcolor{green!10} Qwen3-32B-nothinking & 28.93 & 7.96 & 58.33 & 31.36 & 26.19 & 14.14 & 9.85 & 75.00 & 20.90 & 16.73 & 18.75 & 8.17 & 55.56 & 10.13 & 16.75 & 15.75 & 7.80 & 90.48 & 13.26 & 17.37 \\
   \rowcolor{green!10} Qwen3-32B-thinking & 48.39 & 28.91 & 75.00 & 26.74 & 41.87 & 56.43 & 26.43 & 66.67 & 19.02 & 43.25 & 47.92 & 20.43 & 66.67 & 13.28 & 34.74 & 49.16 & 33.97 & 76.19 & 11.83 & 39.96 \\ \midrule
\rowcolor{yellow!20} \multicolumn{21}{c}{\textbf{Close-source Methods}} \\ \midrule
   \rowcolor{blue!10} GPT-3.5-turbo & 9.42 & 2.78 & 91.67 & 33.15 & 15.21 & 10.18 & 5.77 & 73.33 & 26.10 & 15.00 & 7.03 & 4.30 & 66.67 & 13.29 & 11.83 & 11.64 & 9.14 & 71.43 & 17.72 & 15.29 \\
   \rowcolor{blue!10} GPT-4o & 42.56 & 34.96 & 75.00 & 32.74 & 40.60 & 43.98 & 30.03 & 33.33 & 24.73 & 36.77 & 32.99 & 17.90 & 33.33 & 14.98 & 25.34 & 45.34 & 29.07 & 47.62 & 16.62 & 36.17 \\
   \rowcolor{blue!10} Gemini-2.0 & 39.91 & 33.00 & 75.00 & 28.58 & 38.01 & 29.39 & 33.44 & 53.33 & 18.19 & 28.91 & 27.60 & 27.53 & 44.44 & 5.34 & 23.46 & 29.13 & 29.41 & 38.10 & 14.16 & 26.70 \\
   \rowcolor{blue!10} Gemini-2.5 & 60.45 & 45.37 & 75.00 & 32.94 & 53.37 & 64.84 & 38.28 & 73.33 & 25.57 & 51.70 & 51.69 & 69.57 & 44.44 & 9.73 & 45.41 & 59.35 & 50.87 & 76.19 & 13.59 & 49.40 \\
   \rowcolor{blue!10} Deepseek-Chat-V3 & 50.55 & 22.97 & 75.00 & 38.05 & 43.66 & 56.11 & 21.82 & 53.33 & 29.88 & 43.56 & 55.52 & 15.16 & 66.67 & 11.23 & 36.64 & 48.24 & 21.47 & 66.67 & 18.26 & 37.33 \\ \midrule

\rowcolor{yellow!20} \multicolumn{21}{c}{\textbf{Open-Source Training Methods}} \\ \midrule
   \rowcolor{pink!10} Llama-3.1-8B-Instruct-SFT & 41.20 & 15.35 & 75.00 & 34.15 & 36.07 & 42.50 & 19.33 & 91.67 & 23.56 & 35.74 & 38.02 & 17.29 & 66.67 & 9.73 & 28.53 & 37.11 & 21.50 & 80.95 & 14.89 & 31.64 \\
   \rowcolor{pink!10} Llama-3.1-8B-Instruct-SFT-GRPO & 48.23 & 47.20 & 45.64 & 37.04 & 45.81 & 39.74 & 42.77 & 60.00 & 48.94 & 43.10 & 54.61 & 42.26 & 40.34 & 35.69 & 46.40 & 51.68 & 49.22 & 54.88 & 38.51 & 48.91 \\
   \rowcolor{pink!10} Qwen3-8B-SFT & 48.69 & 28.97 & 75.00 & 37.93 & 43.90 & 51.67 & 28.03 & 40.00 & 24.42 & 40.77 & 44.87 & 20.92 & 66.67 & 13.32 & 33.43 & 44.68 & 22.48 & 61.90 & 17.02 & 35.38 \\
   \rowcolor{pink!10} Qwen3-8B-SFT-GRPO & 51.81 & 41.26 & 42.86 & 31.95 & 45.67 & 49.38 & 28.49 & 71.43 & 36.16 & 43.51 & 48.51 & 24.19 & 55.56 & 17.67 & 36.18 & 42.64 & 27.70 & 46.43 & 32.03 & 37.51 \\
  
\bottomrule
\end{tabular}}
\caption{The main results of current advanced LLMs on \benchmark{} are presented across different language families and tasks. Tasks include Numerical Computation (NC), Cell Extraction (CE), Factual Verification (FV), and Open-Ended Question (OEQ). The average scores (Avg.) across all test cases under certain filter conditions are also presented. For LLMs that include thinking modes, we tested them both with(-thinking) and without(-nothinking) thinking mode enabled.}
\label{tab:Main result}
\end{table*}

\subsection{Experiment Setup}
\label{sec:experimental_setup}
\paragraph{LLMs.} 
We evaluated 20 models encompassing both open-source and proprietary architectures. The open-source models ranged in scale from 7B to 34B parameters, including Baichuan2 \cite{yang2025baichuan2openlargescale}, DeepSeek-Chat \cite{deepseekai2024deepseekllmscalingopensource}, Seed-Coder \cite{seed2025seedcoderletcodemodel}, LLaMA3, LLaMA3.1 \cite{llama3modelcard}, Qwen3 \cite{qwen3technicalreport}, GLM4 \cite{glm2024chatglm}, Phi-4 \cite{abdin2024phi4technicalreport}, Moonlight \cite{liu2025muonscalablellmtraining}, and ERNIE \cite{ernie2025technicalreport}. For proprietary models, we assessed Gemini-2.0, Gemini-2.5 \cite{comanici2025gemini}, GPT-3.5-turbo, and GPT-4o \cite{achiam2023gpt}. Additionally, we performed fine-tuning and reinforcement learning on Qwen3-8B and LLaMA3.1-8B to further investigate LLMs' multilingual table comprehension capabilities.

\begin{table*}[h]
\centering
\resizebox{0.95 \textwidth}{!}{
\begin{tabular}{l|cccccccccccccccccccc}
\toprule
Method & Aar & Af & Sq & Arq & Am & Ar & As & Ban & Bal & Eu & Be & Bn & Bho & Bs & Bul & My & Csx & Ca & Nya & Zh  \\ \midrule
    Qwen3-8B & 13.34 & 31.88 & 27.12 & 21.37 & 23.59 & 33.73 & 28.26 & 29.86 & 25.52 & 38.20 & 31.86 & 30.27 & 32.42 & 26.76 & 28.04 & 16.42 & 24.77 & 33.24 & 16.15 & 30.98 \\ 
    +SFT & 20.59 & 33.29 & 38.72 & 29.20 & 30.33 & 36.13 & 33.79 & 45.65 & 42.25 & 40.84 & 30.99 & 45.68 & 38.94 & 38.25 & 37.57 & 27.77 & 24.43 & 43.17 & 29.69 & 32.35 \\
    +SFT+GRPO & 22.92 & 42.75 & 46.91 & 38.70 & 36.05 & 46.52 & 45.08 & 50.31 & 41.74 & 46.32 & 43.72 & 51.18 & 47.77 & 45.97 & 48.00 & 27.79 & 34.82 & 45.81 & 27.97 & 39.78 \\
    \midrule
Method & Hr & Cs & Da & Nl & En & Et & Fil & Fi & Fr & Ka & De & El & Kal & Gu & Ha & He & Hi & Hu & Is & Id \\ \midrule
    Qwen3-8B & 30.50 & 35.31 & 35.19 & 34.92 & 26.66 & 31.87 & 33.93 & 31.11 & 36.57 & 26.91 & 32.78 & 27.52 & 17.11 & 33.02 & 6.38 & 37.55 & 38.21 & 36.83 & 27.97 & 37.51 \\
    +SFT & 45.36 & 39.48 & 40.77 & 46.45 & 39.79 & 35.69 & 47.72 & 38.67 & 39.53 & 34.84 & 40.54 & 42.44 & 28.31 & 38.71 & 27.43 & 46.25 & 35.56 & 50.07 & 47.08 & 48.54 \\
    +SFT+GRPO & 52.67 & 50.61 & 47.72 & 46.62 & 43.87 & 47.68 & 50.40 & 40.54 & 46.62 & 48.78 & 43.80 & 46.23 & 22.15 & 45.45 & 26.67 & 45.38 & 44.98 & 50.84 & 46.31 & 60.57 \\ \midrule
Method & Ga & It & Ja & Jav & Xal & Kn & Kk & Rw & Kon & Ko & Ky & Lao & Lat & Lv & Li & Mk & Mg & Ms & Ml & Mt \\ \midrule
    Qwen3-8B & 26.16 & 29.96 & 31.52 & 32.47 & 32.14 & 29.53 & 36.48 & 12.87 & 17.42 & 31.89 & 21.01 & 19.30 & 25.46 & 37.72 & 35.55 & 34.84 & 10.46 & 36.20 & 23.79 & 23.41 \\
    +SFT & 45.73 & 38.49 & 36.11 & 37.92 & 45.20 & 38.27 & 47.66 & 21.37 & 33.72 & 39.62 & 36.51 & 30.57 & 42.32 & 44.14 & 47.72 & 40.80 & 25.40 & 41.91 & 38.52 & 36.99 \\
    +SFT+GRPO & 47.79 & 47.25 & 46.33 & 45.72 & 43.57 & 42.10 & 51.20 & 35.47 & 32.11 & 48.78 & 41.83 & 31.99 & 41.99 & 52.49 & 49.53 & 49.04 & 38.02 & 48.84 & 47.56 & 43.66 \\ \midrule
Method & Mri & Mr & Mon & Mos & Ne & No & Ps & Fa & Pl & Pt & Pa & Que & Ro & Rom & Ru & San & Sr & Si & Sk & Sl \\ \midrule
    Qwen3-8B & 10.31 & 20.93 & 29.48 & 7.08 & 27.73 & 38 & 23.74 & 33.84 & 30.85 & 36.49 & 26.40 & 7.51 & 38.37 & 17.72 & 29.36 & 30.10 & 38.07 & 16.77 & 37.24 & 29.16 \\
    +SFT & 31.58 & 34.86 & 36.97 & 13.67 & 35.06 & 38.60 & 37.25 & 45.86 & 38.43 & 42.26 & 45.97 & 27.96 & 42.70 & 36.83 & 37.72 & 37.89 & 43.93 & 32.39 & 47.25 & 40.83 \\
    +SFT+GRPO & 43.52 & 44.26 & 47.66 & 13.61 & 42.46 & 46.65 & 46.00 & 51.86 & 40.38 & 45.06 & 45.97 & 27.40 & 50.15 & 35.10 & 42.99 & 42.45 & 45.48 & 47.53 & 55.81 & 44.72 \\ \midrule
Method & Som & Es & Swa & Sv & Tg & Ta & Te & Th & Tr & Uk & Ur & Ug & Uz & Vi & Xh & Yor & Zu & Avg. & & \\  \midrule
    Qwen3-8B & 12.74 & 39.66 & 21.42 & 30.51 & 18.90 & 31.33 & 26.79 & 27.20 & 37.69 & 32.37 & 37.63 & 17.19 & 34.94 & 28.78 & 11.93 & 18.52 & 13.61 & 28.48 & & \\
    +SFT & 37.76 & 43.82 & 35.98 & 38.00 & 30.56 & 51.78 & 46.06 & 35.62 & 39.18 & 43.64 & 46.47 & 38.56 & 41.52 & 46.25 & 30.21 & 36.33 & 23.49 & 38.56 & & \\
    +SFT+GRPO & 34.29 & 48.79 & 45.46 & 47.84 & 37.88 & 54.48 & 53.55 & 39.39 & 45.90 & 45.04 & 49.95 & 42.80 & 51.99 & 50.39 & 26.51 & 34.71 & 29.30 & 44.15 & & \\
  
\bottomrule
\end{tabular}}
\caption{The performance of the Qwen3-8B model in various languages after SFT and GRPO}
\label{tab:language result}
\end{table*}

\paragraph{Implementation Details.} 
For locally deployable models, we conducted evaluations on an 8×NVIDIA A800 GPU cluster (80GB memory) using the PyTorch Transformer framework. Proprietary models were accessed through official APIs to ensure consistency. During SFT, all model parameters were optimized via the Adam optimizer with a global batch size of 64 achieved through gradient accumulation. Training employed cosine annealing scheduling with an initial learning rate of $1e^{-5}$ over 5 epochs, processing sequences up to 12,000 tokens, which required approximately 20 hours of training time. The GRPO phase utilized a per-device batch size of 2 (aggregating to global batch size of 16 via gradient accumulation), with a constant learning rate of $1e^{-6}$ over 2 epochs. The maximum prompt word length and maximum response length are 12000 and 4096 respectively, and the group number for each question is 5, requiring approximately 36 hours on identical hardware.

\paragraph{Metrics.} 
Consistent with the Task Definition Section, evaluation is performed separately for distinct answer types, with the aggregate score computed as the arithmetic mean across all samples. Due to strict output formatting constraints enforced by prompting, only the answer segment of LLM responses is extracted for evaluation.

\subsection{Main Results}
Table \ref{tab:Main result} presents comprehensive performance metrics of various LLMs across tasks and language families. Among open-source LLMs, the Qwen3 series demonstrated superior overall performance, with Qwen3-32B outperforming counterparts on multiple tasks. For proprietary models, Gemini-2.5 maintained the highest performance level. Regarding task categories, Factual Verification achieved optimal results, partially attributable to its inherent task characteristics and evaluation metrics. As model scale increases, the most substantial improvements are observed in numerical computation tasks. In contrast, most models exhibited comparable performance on Open-Ended Question answering tasks. Across language families, models achieved peak performance on Indo-European languages, while exhibiting the lowest performance on Niger-Congo languages. \ref{fig:language family scores} This performance disparity is likely attributable to variations in corpus quantity and quality across language families in training data sources.

We implemented SFT and GRPO on open-source models Qwen3-8B and Llama-3.1-8B. Results demonstrate that SFT substantially enhances performance across most language families and tasks. The SFT-enhanced Qwen3-8B surpasses GPT-4o in majority scenarios and approaches DeepSeek-V3 capability levels. Concurrently, Llama-3.1-8B exhibits significant improvement reaching GPT-4o parity. These findings validate the efficacy of our training methodology and dataset curation.
Subsequent GRPO training further augmented overall model capability, although minor performance degradation emerged in specific contexts. Overall, GRPO significantly strengthens multilingual table comprehension.
The complete training pipeline yields state-of-the-art performance among open-source LLMs. Notably, the fully-trained Llama-3.1-8B model exceeds DeepSeek-V3 in comprehensive evaluation, ranking second only to Gemini-2.5 among all evaluated systems.

\begin{figure}[ht!]
\begin{center}
    \includegraphics[width=0.45\textwidth]{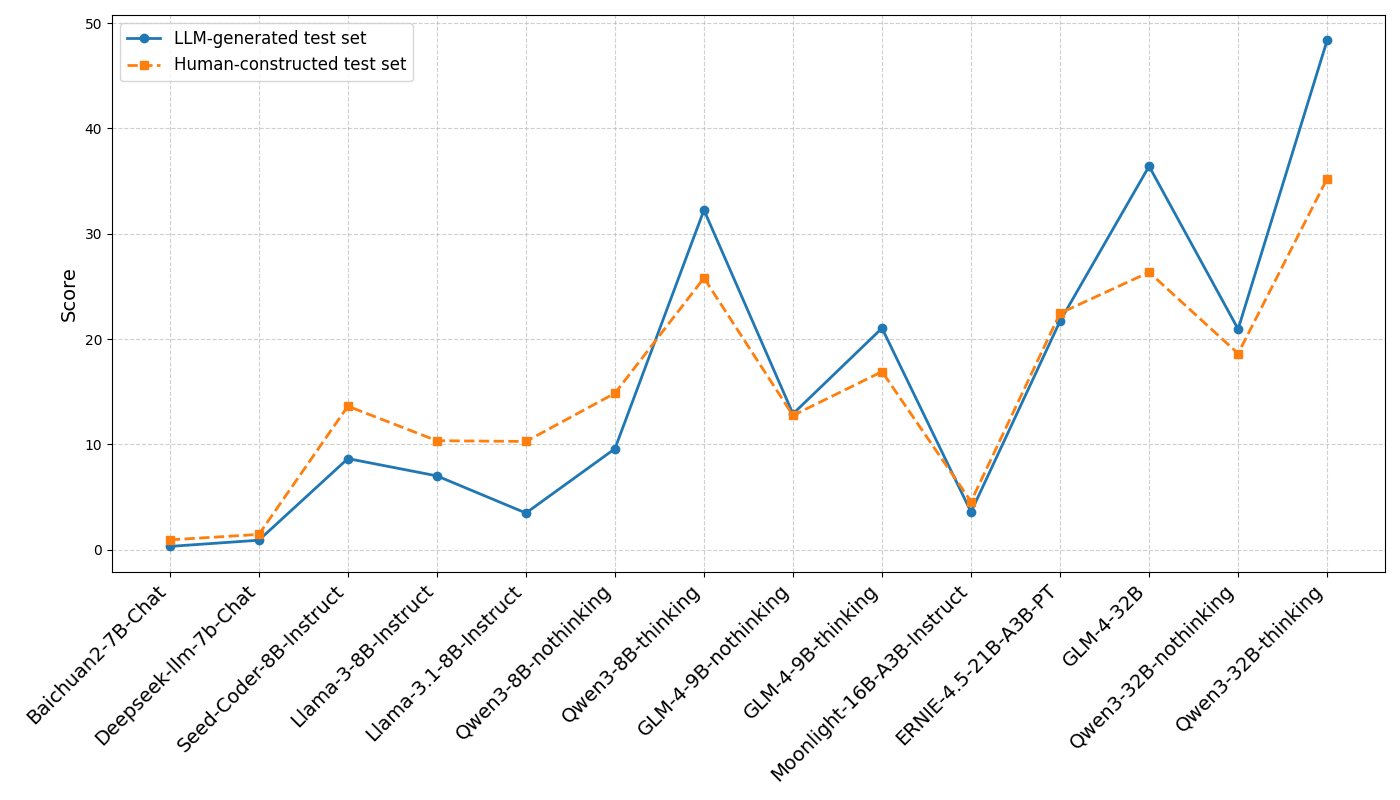}
    \caption{The performance of the LLM on test sets derived from two distinct sources.}
    \label{fig:test compare}
    \vspace{-15pt}
\end{center}
\end{figure}

\textbf{For specific languages: } 
The introduction of SFT produces substantial performance gains in most languages, as evidenced in Table \ref{tab:language result}. The average score improvement reaches 10.46 points in general, with particularly pronounced enhancements exceeding 20 points for low-resource languages, including Maori (Austronesian), Xhosa (Niger-Congo) and Somali (Afro-Asiatic).
Subsequent GRPO training further increases comprehensive performance (from 38.56 to 44.15). However, language-specific analysis reveals divergent outcomes: Performance degradation occurs in Cambodian during SFT, while GRPO induces regression across multiple additional languages.This may be due to the cross-influence between multiple languages in \instruct{}, and indirectly reflects the complexity of multilingual tasks. These limitations underscore the need for more robust cross-lingual generalization techniques.

\textbf{LLMs performance on different test sets: } 
Figure \ref{fig:test compare} presents the performance of different models on the two source test sets. Scores on both test sets exhibit an upward trend as model parameter count increases. However, a distinct pattern emerges: smaller-scale models (below 16B parameters) generally achieve superior performance on the human-curated test set, whereas larger-scale models (above 16B parameters) tend to perform better on the LLM-generated test set. Furthermore, enabling the model's thinking mode significantly enhance the model's performance on the LLM-generated test set. These results delineate distinct performance advantage zones associated with model scale and setting. It underscores the necessity of utilizing dual test sets, as a single benchmark is insufficient for comprehensively evaluating all facets of model capability. Consequently, this dual-test-set approach renders the evaluation leaderboard more comprehensive and challenging.

\begin{figure}[h]
\begin{center}
    \includegraphics[width=0.45\textwidth]{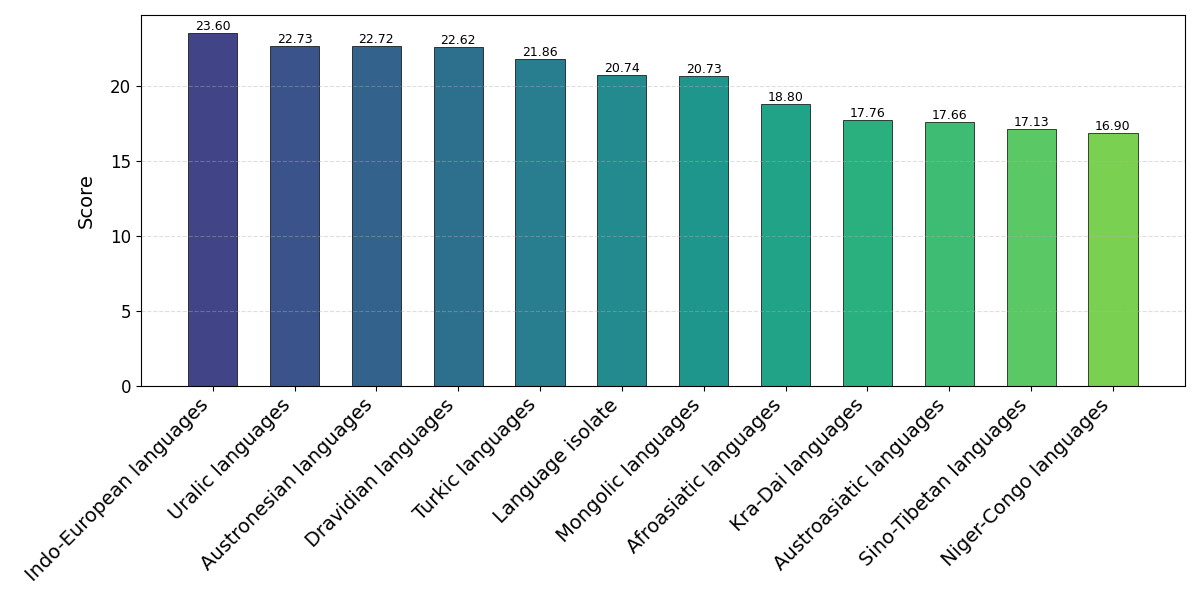}
    \caption{Average score for each language family}
    \label{fig:language family scores}
    \vspace{-15pt}
\end{center}
\end{figure}

\subsection{Ablation study}
We employed Qwen3-8B as the base model for ablation studies to examine the impact of thinking mode, SFT, and GRPO on cross-lingual performance. As shown in Table \ref{tab:Ablation study}, both SFT and enabling the thinking mode yielded substantial and comparable performance improvements over the base model. Specifically, SFT achieved a relative improvement of 119.75\%, while activating the thinking mode resulted in a 124.96\% enhancement. This demonstrates that both interventions provide similarly significant gains, with thinking mode exhibiting a marginally higher uplift. Following the completion of SFT, we further optimized the model using the GRPO algorithm. Experimental results demonstrate that this optimization yielded a comprehensive performance improvement of approximately 6 points across both the thinking mode and non-thinking mode settings. At the language family level, GRPO delivered marked improvements exceeding 10 points for Afroasiatic, Sino-Tibetan, and Niger-Congo language families while generating measurable gains for most other families. 
Notably, however, performance degradation was observed in the Austronesian and Austroasiatic language families. 

In conclusion, the whole training pipeline achieved optimal performance on \benchmark{}. Crucially, the systematic incorporation of thinking capabilities yielded significant incremental gains at different training stages: Base pre-training (15.82 points), SFT (10.74 points), and GRPO (10.49 points). These results demonstrate its essential role in enhancing complex multilingual table understanding.

\begin{table}[ht]
\centering
\resizebox{0.45 \textwidth}{!}{
\begin{tabular}{lllccccccc}
\toprule
\multicolumn{3}{l}{Language family} & Afroasiatic & Indo-European & Austronesian & Sino-Tibetan &
Austroasiatic & Niger-Congo & Uralic  \\ \midrule
    \multicolumn{3}{l}{Qwen3-8B} & 12.72 & 13.22 & 14.11 & 6.74 & 11.79 & 12.28 & 9.29 \\
    & \multicolumn{2}{l}{w/ SFT} & 23.43 & 29.97 & 28.12 & 22.94 & 12.65 & 21.43 & 26.86 \\
    & & w/ GRPO & 31.71 & 35.87 & 25.77 & 33.41 & 24.61 & 33.07 & 31.84 \\
    \multicolumn{3}{l}{w/ Thinking} & 22.40 & 30.84 & 27.71 & 24.30 & 24.77 & 15.75 & 33.24 \\
    & \multicolumn{2}{l}{w/ SFT} & 33.69 & 40.41 & 40.10 & 30.25 & 36.60 & 28.66 & 41.35 \\
    & & w/ GRPO & 46.41 & 46.27 & 34.28 & 48.49 & 31.58 & 46.94 & 41.98 \\ \midrule
\multicolumn{3}{l}{Language family} & Mongolic & Dravidian & Turkic & Kra-Dai & Language isolate & Avg. & \\ \midrule
    \multicolumn{3}{l}{Qwen3-8B} & 12.72 & 13.52 & 10.57 & 11.74 & 12.04 & 12.66 & \\
    & \multicolumn{2}{l}{w/ SFT} & 28.88 & 29.03 & 27.79 & 19.60 & 24.99 & 27.82 & \\
    & & w/ GRPO & 39.08 & 36.42 & 34.69 & 30.69 & 28.74 & 33.66 & \\
    \multicolumn{3}{l}{w/ Thinking} & 30.77 & 27.86 & 29.67 & 23.78 & 27.23 & 28.48 & \\
    & \multicolumn{2}{l}{w/ SFT} & 40.97 & 43.90 & 40.77 & 33.43 & 35.38 & 38.56 & \\
    & & w/ GRPO & 49.69 & 45.67 & 43.51 & 36.18 & 37.51 & 44.15 & \\ 
\bottomrule
\end{tabular}}
\caption{The impact of thinking mode, SFT, and GRPO.}
\label{tab:Ablation study}
\end{table}

\section{Analysis}
\paragraph{Effectiveness of training data generated by LLM}
We observe that despite \instruct{} being entirely LLM-generated without human curation, models fine-tuned and reinforced with this dataset exhibit substantial improvement on purely human-constructed test sets. This finding potentially informs LLM optimization strategies, suggesting that task-relevant synthetic training data generated by LLMs can effectively enhance base model performance at lower operational costs.

\paragraph{Impact of language family on LLM reasoning}
Our analysis across language families (Fig. \ref{fig:language family scores}) reveals that Indo-European languages achieve the highest scores. This advantage stems from abundant high-quality corpora (e.g., English) produced by its extensive and widely distributed speaker base, whose generated data effectively supports LLM learning.
However, the Niger-Congo language family - although it has 800 million native speakers predominantly in underdeveloped regions - yields limited corpora, reflected in lower LLM performance. In contrast, Uralic languages (25 million speakers) exhibit comparatively stronger performance. This disparity arises from Uralic speakers' concentration in developed regions, generating disproportionately greater digital resources than Niger-Congo languages. For example, Finnish (Uralic) has 598,717 Wikipedia articles versus Swahili's (Niger-Congo) 100,179 articles. This resource asymmetry suggests researcher bias toward digitally prevalent languages regardless of speaker population size, highlighting significant imbalances in linguistic resource allocation that challenge equitable LLM development.
\begin{figure}[h]
\begin{center}
    \includegraphics[width=0.45\textwidth]{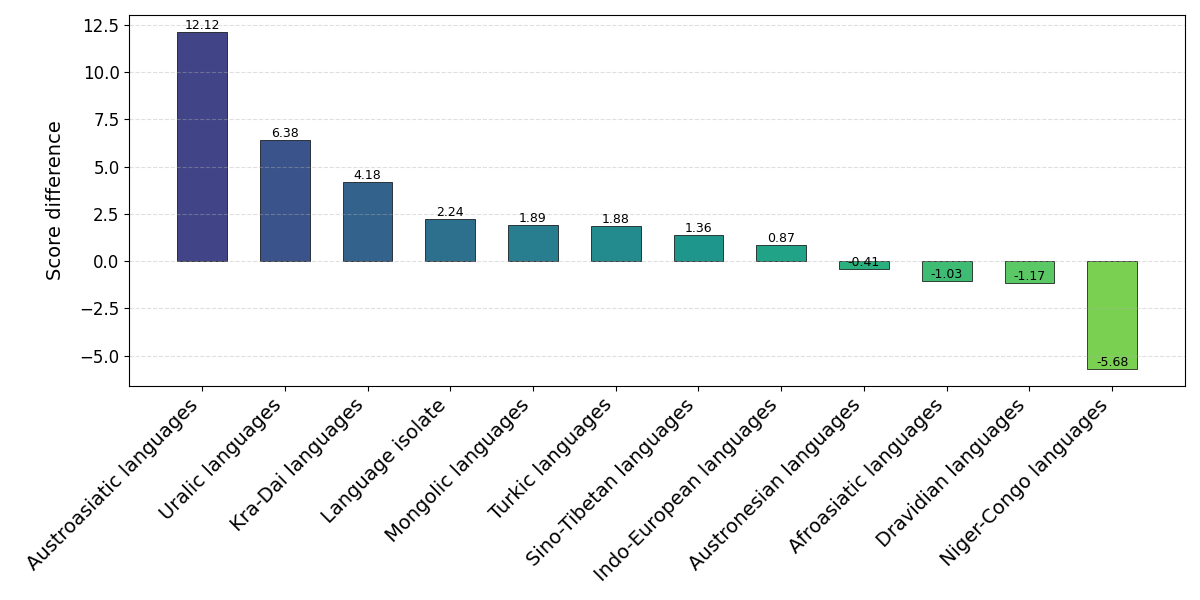}
    \caption{Difference between thinking mode and SFT score (thinking mode score - SFT score)}
    \label{fig:4}
\end{center}
\end{figure}

\paragraph{Thinking mode and SFT}
Analysis of Qwen3-8B reveals distinct cross-linguistic response patterns to thinking mode versus SFT, despite comparable overall performance gains. We identify a pronounced geographical correlation: Thinking mode yields superior enhancement for languages predominantly spoken in developed regions (Indo-European, Sino-Tibetan, Uralic), while SFT demonstrates greater efficacy for languages concentrated in underdeveloped regions (Niger-Congo, Afro-Asiatic, Austronesian)(Fig. \ref{fig:4}). This divergence reflects fundamental disparities in linguistic resource availability within LLMs' knowledge bases. For digitally prevalent languages, inherent high-quality representations enable effective task resolution through structured reasoning alone. Conversely, under-resourced languages benefit disproportionately from SFT's capacity to mitigate knowledge gaps through task-aligned external supervision.

\section{Related Work}
\paragraph{Development of Table QA Datasets.} 
To advance research on Table Question Answering (TQA), several English benchmark datasets have been proposed~\citep{akhtar2025tanq, chen2020hybridqa, zhong2017seq2sql, pasupat2015compositional}. These datasets play a crucial role in evaluating models' semantic understanding capabilities regarding tabular structures. Currently, benchmarks such as~\cite{chang2023dr, zhao2023robut, sui2024table, wu2025tablebench} specifically assess large language models' (LLM) table comprehension abilities~\citep{tiwari2025auto, ye2023large, wang2024chain, wu2025table, zheng2024multimodal, titiya2025mmtbench}. However, these datasets remain exclusively English-centric, failing to meet the evaluation needs of modern multilingual LLMs.

\paragraph{Exploration of Multilingual Datasets.}
To address the multilingual evaluation gap, researchers have introduced non-English QA datasets~\citep{liu2019xqa, clark2020tydi}. OMGEval~\citep{liu2024omgeval} established an open-domain evaluation framework that covers five languages,ges, including Chinese, French, and Russian. In the tabular data domain, IM-TQA~\citep{zheng2023tqa} constructed benchmarks for table classification and QA using Chinese corporate annual reports and encyclopedic web pages. ~\cite{zhang2025multitat, minhas2022xinfotabs} expanded existing TQA datasets to 10 languages (covering ~3 billion native speakers) through translation techniques, enabling cross-language family and cross-script evaluation. For low-resource languages, recent studies~\citep{cho2025multilingual, pal2024table} addressed data scarcity through selective question translation and cross-lingual adaptation methods, while analyzing the impact of linguistic features (e.g., word order) on model performance.

\section{Conclusion}
In this work, we introduce \benchmark{}, a comprehensive and linguistically complex table question-answering dataset to evaluate multilingual tabular reasoning capabilities. Spanning 97 languages across 12 language families, \benchmark{} contains 2,916 LLM-generated QA pairs and 4,294 human-curated QA pairs with meticulous annotations. 
We benchmarked 20+ models on \benchmark{} to quantify cross-lingual performance variations under diverse task constraints. To enhance multilingual adaptability, we create \instruct{} to provide a resource-efficient solution for multilingual optimization. Empirical results demonstrate that models trained on \instruct{} achieve significant performance gains despite the absence of human annotation. \benchmark{} establishes a new evaluation standard for multilingual table understanding and delivers a rigorous benchmarking platform for assessing model robustness across linguistic typologies.


\bigskip

\bibliography{aaai2026}
\clearpage


\appendix

\onecolumn 
\section{LLM Prompt Examples}
Figure 7~10 show the prompt for different tasks, where \texttt{\{Table data\}} is the serialized table data and \texttt{\{Question\}} is the table question.
\begin{figure}[ht!]
\begin{center}
    \includegraphics[width=0.55\textwidth]{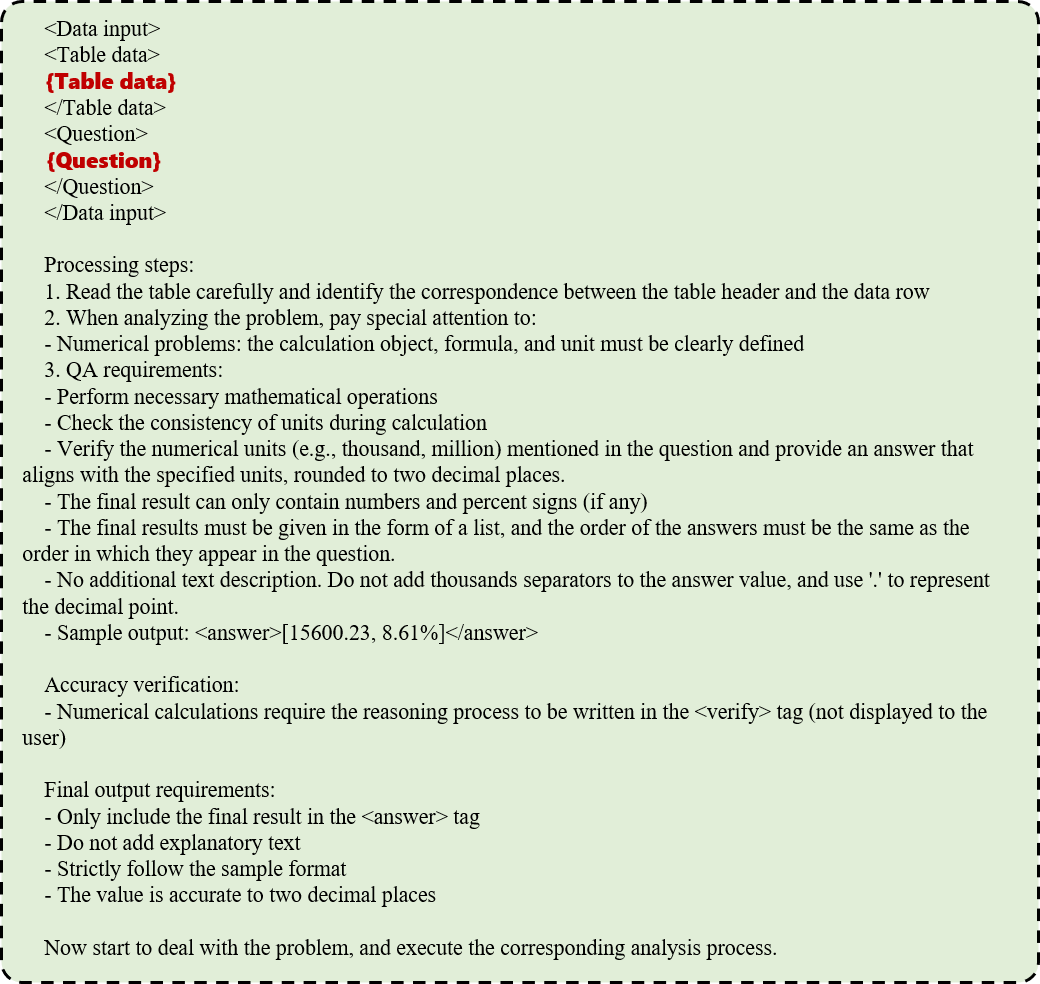}
    \caption{Instruction of Numerical Computation}
\end{center}
\end{figure}

\begin{figure}[ht!]
\begin{center}
    \includegraphics[width=0.55\textwidth]{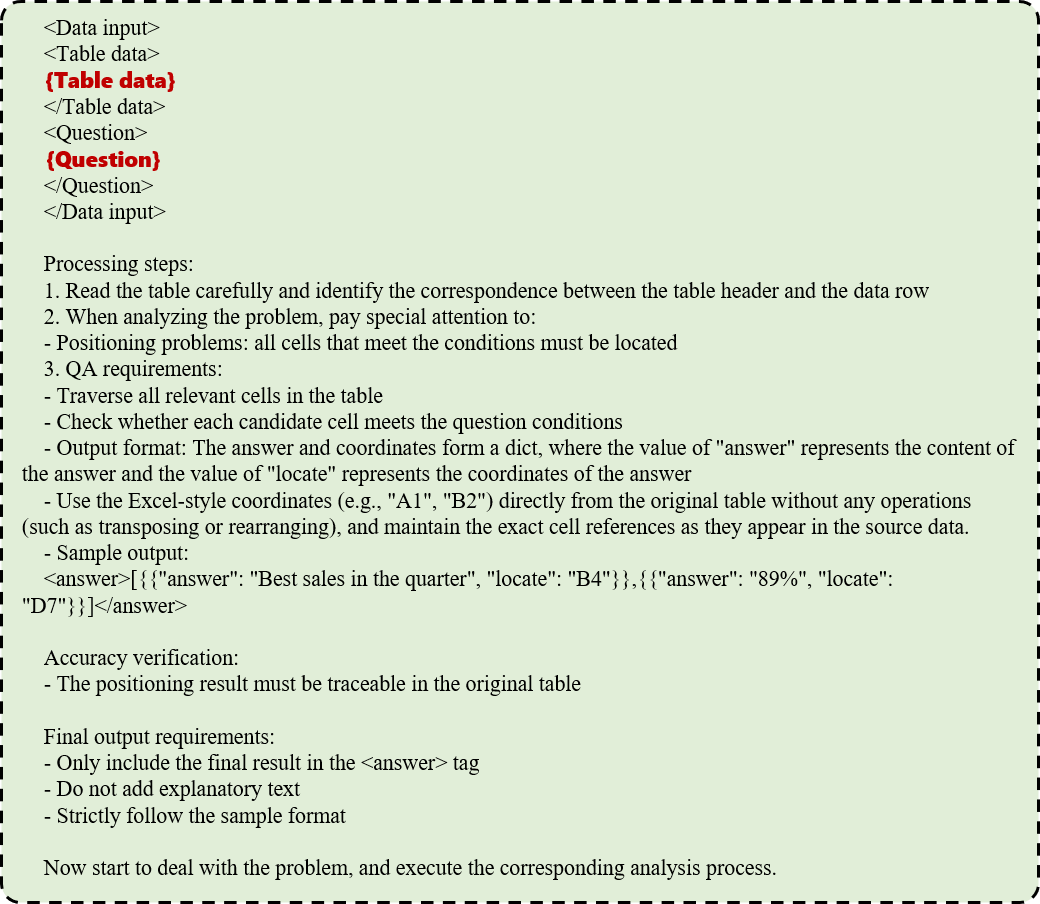}
    \caption{Instruction of Cell Extraction}
\end{center}
\end{figure}

\begin{figure}[ht!]
\begin{center}
    \includegraphics[width=0.55\textwidth]{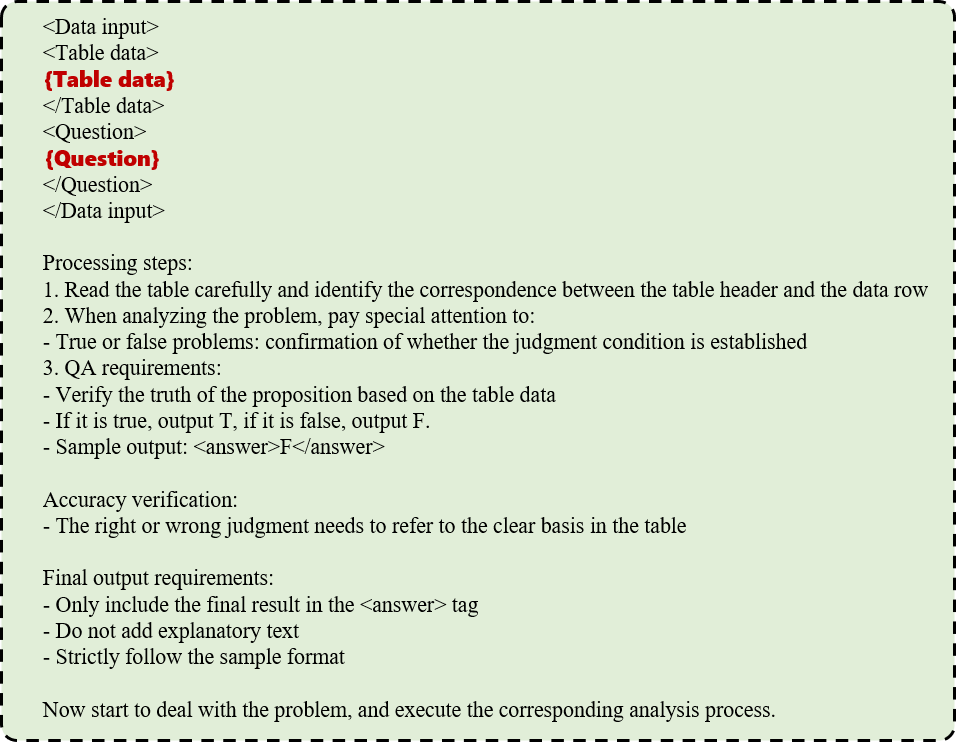}
    \caption{Instruction of Factual Verification}
\end{center}
\end{figure}

\begin{figure}[ht!]
\begin{center}
    \includegraphics[width=0.55\textwidth]{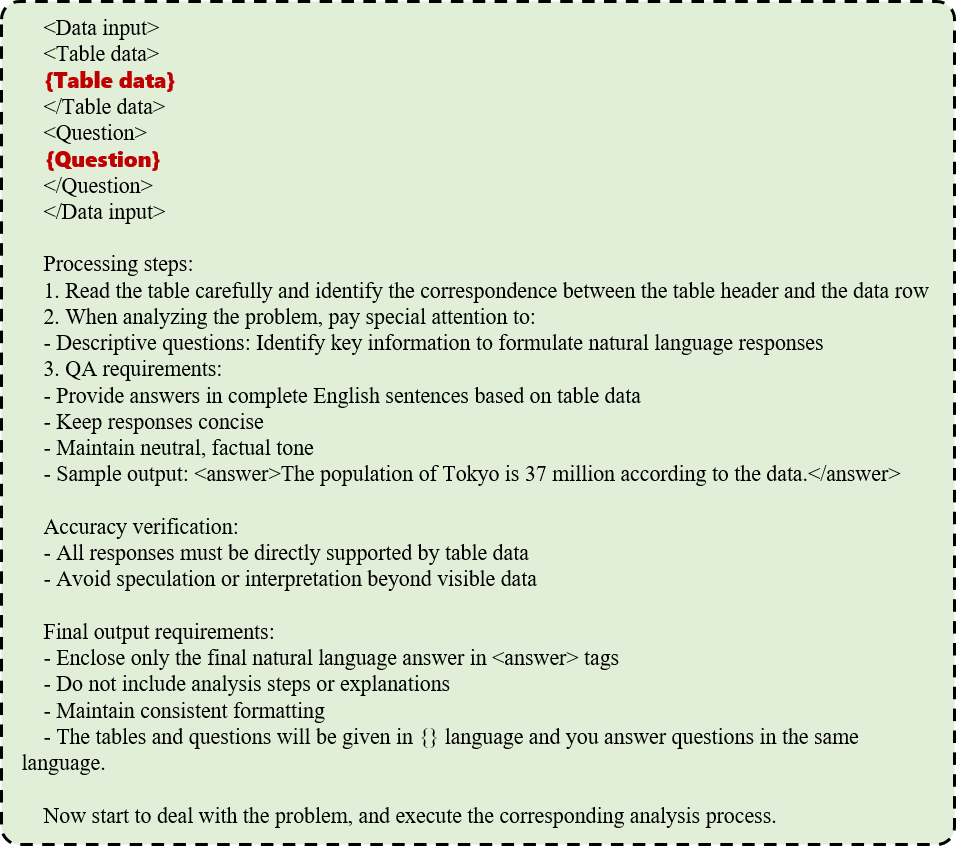}
    \caption{Instruction of Open-Ended Questions}
\end{center}
\end{figure}

\twocolumn 

\section{Statistics for all languages}\label{app:Data Statistics}
Table \ref{tab:language_stats} summarizes the dataset by language, including the language family to which each language belongs and the amount of training and test data it has.

\section{Results of different types of tables}\label{app:Results of different types of tables}
Table \ref{tab:Table type result} shows the performance of different models on each type of table. Overall, most models perform better on entity tables, which may be related to the horizontal arrangement of information.
\begin{table}[ht]
\centering
\resizebox{0.45 \textwidth}{!}{
\begin{tabular}{l|cccc}
\toprule
& Relational table & Entity table & Matrix table & Composite table  \\ \midrule
\multicolumn{5}{c}{\textbf{Open-source Methods}} \\ \midrule
   Baichuan2-7B-Chat & 0.57 & 0.55 & 0.81 & 0.79  \\ 
   Deepseek-llm-7B-Chat & 0.96 & 2.16 & 1.02 & 1.66  \\
   Seed-Coder-8B-Instruct & 13.15 & 11.46 & 8.70 & 11.99 \\ 
   Llama-3-8B-Instruct & 7.27 & 15.97 & 8.78 & 12.85 \\
   Llama-3.1-8B-Instruct & 6.43 & 15.77 & 8.11 & 11.35 \\
   Qwen3-8B-nothinking & 13.13 & 12.40 & 10.13 & 17.07 \\
   Qwen3-8B-thinking & 26.45 & 41.34 & 27.10 & 26.45 \\
   GLM-4-9B-0414-nothinking & 10.78 & 19.98 & 11.71 & 15.96 \\
   GLM-4-9B-0414-thinking & 17.53 & 29.15 & 15.19 & 20.11 \\
   Phi-4-14B & 25.32 & 35.67 & 22.72 & 25.74 \\
   Moonlight-16B-A3B-Instruct & 4.09 & 5.52 & 3.50 & 4.59 \\
   ERNIE-4.5-21B-A3B-PT & 22.47 & 24.50 & 19.72 & 24.10 \\
   GLM-4-32B & 29.83 & 42.10 & 28.15 & 25.99 \\
   Qwen3-32B-nothinking & 19.95 & 20.65 & 18.56 & 19.32 \\
   Qwen3-32B-thinking & 40.62 & 48.68 & 40.85 & 32.42 \\ \midrule
\multicolumn{5}{c}{\textbf{Close-source Methods}} \\ \midrule
   GPT-3.5-turbo & 16.78 & 23.38 & 12.10 & 18.16 \\
   GPT-4o & 35.21 & 45.92 & 32.60 & 33.79 \\
   Gemini-2.0 & 29.72 & 45.23 & 23.90 & 26.27 \\
   Gemini-2.5 & 54.06 & 51.99 & 49.92 & 39.67 \\
   Deepseek-Chat-V3 & 41.70 & 50.88 & 41.55 & 35.71 \\ \midrule
\multicolumn{5}{c}{\textbf{Open-Source Training Methods}} \\ \midrule
   Llama-3.1-8B-Instruct-sft & 34.37 & 43.74 & 33.05 & 31.62 \\
   Qwen3-8B-sft & 37.52 & 46.69 & 38.41 & 34.64 \\
\bottomrule
\end{tabular}}
\caption{Performance of each model on different types of tables.}
\label{tab:Table type result}
\vspace{-10pt}
\end{table}

\begin{table*}[t]
\centering
\small
\resizebox{0.95 \textwidth}{!}{
\begin{tabular}{@{}lllrrlllrr@{}}
\toprule
Code & Language & Language Family & Train & Test & Code & Language & Language Family & Train & Test \\
\midrule
aar & Afar & Afroasiatic & 369 & 60 & ky & Kyrgyz & Turkic & 405 & 80 \\
af & Afrikaans & Indo-European & 423 & 80 & lao & Lao & Kra-Dai & 349 & 59 \\
sq & Albanian & Indo-European & 395 & 84 & lat & Latin & Indo-European & 404 & 73 \\
arq & Algerian Arabic & Afroasiatic & 414 & 76 & lv & Latvian & Indo-European & 422 & 75 \\
am & Amharic & Afroasiatic & 367 & 71 & lt & Lithuanian & Indo-European & 423 & 76 \\
ar & Arabic & Afroasiatic & 431 & 82 & mk & Macedonian & Indo-European & 414 & 82 \\
as & Assamese & Indo-European & 412 & 76 & mg & Malagasy & Austronesian & 369 & 72 \\
ban & Balinese & Austronesian & 422 & 69 & ms & Malay & Austronesian & 404 & 75 \\
bal & Baluchi & Indo-European & 405 & 69 & ml & Malayalam & Dravidian & 411 & 75 \\
eu & Basque & isolate & 405 & 73 & mt & Maltese & Afroasiatic & 404 & 70 \\
be & Belarusian & Indo-European & 404 & 71 & mri & Maori & Austronesian & 351 & 63 \\
bn & Bengali & Indo-European & 413 & 80 & mr & Marathi & Indo-European & 430 & 81 \\
bho & Bhojpuri & Indo-European & 423 & 78 & mon & Mongolian & Mongolic & 404 & 74 \\
bs & Bosnian & Indo-European & 441 & 84 & mos & Mossi & Niger-Congo & 270 & 45 \\
bul & Bulgarian & Indo-European & 414 & 87 & ne & Nepali & Indo-European & 421 & 86 \\
my & Burmese & Sino-Tibetan & 396 & 72 & no & Norwegian & Indo-European & 441 & 79 \\
csx & Cambodian & Austroasiatic & 349 & 65 & ps & Pashto & Indo-European & 386 & 69 \\
ca & Catalan & Indo-European & 414 & 78 & fa & Persian & Indo-European & 403 & 70 \\
nya & Chichewa & Niger-Congo & 387 & 60 & pl & Polish & Indo-European & 423 & 77 \\
zh & Chinese & Sino-Tibetan & 448 & 85 & pt & Portuguese & Indo-European & 423 & 77 \\
hr & Croatian & Indo-European & 422 & 89 & pa & Punjabi & Indo-European & 411 & 70 \\
cs & Czech & Indo-European & 405 & 80 & que & Quechua & isolate & 306 & 50 \\
da & Danish & Indo-European & 405 & 79 & ro & Romanian & Indo-European & 414 & 72 \\
nl & Dutch & Indo-European & 422 & 77 & rom & Romany & Indo-European & 414 & 69 \\
en & English & Indo-European & 432 & 87 & ru & Russian & Indo-European & 413 & 80 \\
et & Estonian & Uralic & 423 & 86 & san & Sanskrit & Indo-European & 439 & 77 \\
fil & Filipino & Austronesian & 404 & 72 & sr & Serbian & Indo-European & 413 & 81 \\
fi & Finnish & Uralic & 423 & 79 & si & Sinhalese & Indo-European & 404 & 73 \\
fr & French & Indo-European & 421 & 82 & sk & Slovak & Indo-European & 405 & 75 \\
ka & Georgian & isolate & 423 & 78 & sl & Slovenian & Indo-European & 432 & 88 \\
de & German & Indo-European & 423 & 77 & som & Somali & Afroasiatic & 405 & 70 \\
el & Greek & Indo-European & 396 & 74 & es & Spanish & Indo-European & 423 & 80 \\
kal & Greenlandic & isolate & 288 & 50 & swa & Swahili & Niger-Congo & 394 & 79 \\
gu & Gujarati & Indo-European & 405 & 77 & sv & Swedish & Indo-European & 414 & 71 \\
ha & Hausa & Afroasiatic & 396 & 67 & tg & Tajik & Indo-European & 369 & 65 \\
he & Hebrew & Afroasiatic & 405 & 85 & ta & Tamil & Dravidian & 405 & 79 \\
hi & Hindi & Indo-European & 403 & 72 & te & Telugu & Dravidian & 414 & 79 \\
hu & Hungarian & Uralic & 405 & 80 & th & Thai & Kra-Dai & 422 & 77 \\
is & Icelandic & Indo-European & 432 & 91 & tr & Turkish & Turkic & 423 & 74 \\
id & Indonesian & Austronesian & 423 & 80 & uk & Ukrainian & Indo-European & 423 & 82 \\
ga & Irish & Indo-European & 396 & 76 & ur & Urdu & Indo-European & 422 & 77 \\
it & Italian & Indo-European & 432 & 87 & ug & Uyghur & Turkic & 359 & 57 \\
ja & Japanese & isolate & 431 & 90 & uz & Uzbek & Turkic & 432 & 79 \\
jav & Javanese & Austronesian & 404 & 67 & vi & Vietnamese & Austroasiatic & 429 & 82 \\
xal & Kalmyk & Mongolic & 395 & 70 & xh & Xhosa & Niger-Congo & 394 & 68 \\
kn & Kannada & Dravidian & 396 & 69 & yor & Yoruba & Niger-Congo & 368 & 59 \\
kk & Kazakh & Turkic & 423 & 76 & zu & Zulu & Niger-Congo & 395 & 68 \\
rw & Kinyarwanda & Niger-Congo & 340 & 56 &  &  &  &  &  \\
kon & Kongo & Niger-Congo & 288 & 47 &  &  &  &  &  \\
ko & Korean & isolate & 385 & 71 &  &  &  &  &  \\
\bottomrule
\end{tabular}}
\caption{Statistics of the training, and test sets for all languages. The languages are ranked in alphabet.}
\label{tab:language_stats}
\end{table*}

\section{All Language Results} \label{app:All Language Results}
Table \ref{tab:all language result} shows the scores of all models on each language. Including test results of open source models, closed source models, and models trained by us.



\onecolumn 
{\scriptsize 
\setlength{\tabcolsep}{0.6mm}{
\begin{longtable}{l|cccccccccccccccccccc}

\hline
\endfirsthead 

\hline
\endhead 

\hline 
\multicolumn{21}{c}{}\\
\multicolumn{21}{c}{\normalsize\tablename\ \thetable\ The performance of all models in various languages.}\\
\endfoot

\hline 
\multicolumn{21}{c}{}\\
\caption{The performance of all models in various languages}
\label{tab:all language result}
\endlastfoot
Method & Aar & Af & Sq & Arq & Am & Ar & As & Ban & Bal & Eu & Be & Bn & Bho & Bs & Bul & My & Csx & Ca & Nya & Zh  \\ \midrule
\multicolumn{21}{c}{\textbf{Open-source Methods}} \\ \midrule
    Baichuan2-7B-Chat & 0.00 & 0.22 & 0.00 & 0.22 & 0.00 & 0.00 & 3.02 & 1.99 & 0.17 & 0.00 & 0.17 & 1.70 & 0.15 & 1.71 & 0.32 & 0.00 & 0.08 & 0.31 & 0.64 & 1.58 \\
    Deepseek-llm-7B-Chat & 0.86 & 2.86 & 0.69 & 3.17 & 0.00 & 0.00 & 0.00 & 0.42 & 0.80 & 1.21 & 1.50 & 2.50 & 2.40 & 2.06 & 1.15 & 0.69 & 0.00 & 0.19 & 0.26 & 1.57 \\
    Seed-Coder-8B-Instruct & 11.87 & 13.90 & 9.69 & 8.09 & 4.31 & 11.85 & 7.77 & 17.71 & 9.81 & 8.74 & 11.27 & 10.15 & 8.63 & 11.78 & 14.65 & 7.32 & 5.78 & 15.08 & 8.39 & 15.10 \\ 
    Llama-3-8B-Instruct & 11.51 & 13.53 & 10.01 & 7.83 & 4.34 & 9.06 & 4.78 & 12.91 & 11.44 & 5.90 & 7.63 & 7.23 & 8.84 & 10.70 & 11.65 & 3.84 & 1.33 & 11.43 & 7.01 & 7.45 \\ 
    Llama-3.1-8B-Instruct & 6.12 & 13.70 & 6.21 & 5.27 & 1.16 & 7.75 & 7.53 & 13.46 & 10.89 & 6.83 & 4.94 & 12.60 & 8.28 & 11.14 & 7.09 & 6.11 & 3.86 & 13.29 & 10.07 & 2.75 \\ 
    Qwen3-8B-nothinking & 16.14 & 14.40 & 10.95 & 11.92 & 9.92 & 11.00 & 14.30 & 12.26 & 12.12 & 9.40 & 10.93 & 13.40 & 13.99 & 9.64 & 12.77 & 8.86 & 11.79 & 13.95 & 10.53 & 4.71 \\ 
    Qwen3-8B-thinking & 13.34 & 31.88 & 27.12 & 21.37 & 23.59 & 33.73 & 28.26 & 29.86 & 25.52 & 38.20 & 31.86 & 30.27 & 32.42 & 26.76 & 28.04 & 16.42 & 24.77 & 33.24 & 16.15 & 30.98 \\ 
    GLM-4-9B-0414-nothinking & 6.54 & 19.03 & 12.77 & 8.86 & 7.46 & 16.72 & 8.52 & 11.95 & 8.05 & 10.84 & 14.56 & 12.86 & 12.31 & 17.07 & 12.40 & 5.46 & 3.85 & 13.02 & 8.45 & 11.96 \\ 
    GLM-4-9B-0414-thinking & 15.34 & 20.50 & 17.82 & 19.00 & 4.03 & 19.58 & 15.03 & 15.28 & 17.72 & 22.49 & 20.99 & 17.53 & 18.89 & 23.94 & 24.17 & 5.30 & 8.52 & 20.09 & 7.53 & 21.57 \\ 
    Phi-4-14B & 17.44 & 26.05 & 34.33 & 18.14 & 9.74 & 27.72 & 27.76 & 25.23 & 30.09 & 24.63 & 21.76 & 26.41 & 27.94 & 26.77 & 30.70 & 13.42 & 14.61 & 32.21 & 13.25 & 21.96 \\ 
    Moonlight-16B-A3B-Instruct & 8.43 & 4.72 & 3.09 & 4.19 & 1.31 & 2.67 & 2.94 & 1.86 & 3.44 & 6.38 & 4.19 & 1.23 & 5.25 & 7.48 & 3.13 & 1.79 & 0.34 & 5.52 & 4.22 & 5.04 \\ 
    ERNIE-4.5-21B-A3B-PT & 13.18 & 24.44 & 20.97 & 13.36 & 17.56 & 23.30 & 21.70 & 27.32 & 20.72 & 21.44 & 25.10 & 20.96 & 23.18 & 21.87 & 20.79 & 16.68 & 17.04 & 19.22 & 28.57 & 18.04 \\ 
    GLM-4-32B & 15.57 & 34.68 & 37.18 & 24.96 & 23.36 & 30.85 & 24.51 & 32.16 & 17.76 & 36.42 & 23.85 & 31.54 & 30.31 & 33.87 & 35.01 & 17.77 & 19.52 & 35.66 & 21.24 & 32.82 \\ 
    Qwen3-32B-nothinking & 19.45 & 15.03 & 25.07 & 15.85 & 19.51 & 16.91 & 23.03 & 16.99 & 17.67 & 18.33 & 18.80 & 25.85 & 27.71 & 22.38 & 15.48 & 16.98 & 15.29 & 19.40 & 13.48 & 11.61 \\ 
    Qwen3-32B-thinking & 22.77 & 50.30 & 50.02 & 33.72 & 34.35 & 48.69 & 30.94 & 45.10 & 35.73 & 45.42 & 37.77 & 44.87 & 44.21 & 33.70 & 46.95 & 32.33 & 34.85 & 50.01 & 18.01 & 38.04 \\ \midrule
\multicolumn{21}{c}{\textbf{Close-source Methods}} \\ \midrule
    GPT-3.5-turbo & 16.17 & 21.58 & 18.73 & 19.29 & 9.64 & 19.28 & 9.73 & 17.78 & 11.04 & 16.78 & 13.76 & 14.63 & 16.15 & 20.67 & 19.82 & 11.27 & 8.96 & 17.92 & 12.21 & 11.02 \\ 
    GPT-4o & 23.38 & 37.36 & 37.56 & 27.58 & 25.27 & 31.86 & 31.67 & 33.03 & 30.19 & 43.29 & 34.18 & 31.14 & 38.00 & 34.65 & 40.70 & 18.96 & 28.14 & 41.01 & 29.09 & 28.01 \\ 
    Gemini-2.0 & 24.34 & 28.60 & 35.62 & 30.59 & 30.32 & 25.91 & 24.57 & 41.72 & 29.90 & 27.24 & 30.84 & 21.37 & 38.26 & 24.42 & 27.25 & 23.49 & 22.12 & 31.12 & 31.30 & 31.57 \\ 
    Gemini-2.5 & 32.46 & 47.82 & 56.57 & 37.12 & 44.92 & 52.00 & 44.03 & 46.48 & 54.38 & 59.32 & 52.38 & 50.89 & 52.48 & 49.28 & 51.70 & 34.95 & 38.61 & 54.67 & 47.43 & 52.35 \\ 
    Deepseek-Chat-V3 & 30.18 & 39.21 & 49.50 & 30.92 & 31.14 & 45.21 & 36.21 & 41.80 & 41.75 & 38.65 & 44.93 & 49.97 & 42.09 & 40.15 & 44.48 & 31.14 & 27.84 & 45.63 & 34.49 & 37.35 \\ \midrule
\multicolumn{21}{c}{\textbf{Open-Source Training Methods}} \\ \midrule
    Llama-3.1-8B-Instruct-SFT & 25.31 & 33.49 & 37.79 & 27.92 & 18.50 & 34.43 & 40.27 & 38.75 & 39.39 & 36.19 & 30.00 & 35.71 & 34.26 & 35.25 & 40.89 & 22.47 & 18.15 & 39.16 & 26.87 & 25.39 \\ 
    Llama-3.1-8B-Instruct-SFT-GRPO & 49.40 & 47.35 & 50.82 & 43.35 & 45.33 & 52.72 & 49.82 & 45.39 & 49.64 & 58.40 & 49.73 & 53.61 & 42.67 & 46.25 & 58.91 & 42.42 & 41.54 & 49.61 & 42.94 & 45.67 \\ 
    
    Qwen3-8B-SFT & 20.59 & 33.29 & 38.72 & 29.20 & 30.33 & 36.13 & 33.79 & 45.65 & 42.25 & 40.84 & 30.99 & 45.68 & 38.94 & 38.25 & 37.57 & 27.77 & 24.43 & 43.17 & 29.69 & 32.35 \\
    Qwen3-8B-SFT-GRPO & 22.92 & 42.75 & 46.91 & 38.70 & 36.05 & 46.52 & 45.08 & 50.31 & 41.74 & 46.32 & 43.72 & 51.18 & 47.77 & 45.97 & 48.00 & 27.79 & 34.82 & 45.81 & 27.97 & 39.78 \\
    \midrule
Method & Hr & Cs & Da & Nl & En & Et & Fil & Fi & Fr & Ka & De & El & Kal & Gu & Ha & He & Hi & Hu & Is & Id \\ \midrule
\multicolumn{21}{c}{\textbf{Open-source Methods}} \\ \midrule
    Baichuan2-7B-Chat & 0.00 & 0.68 & 0.27 & 1.27 & 3.05 & 0.00 & 1.16 & 0.21 & 0.00 & 0.05 & 1.88 & 0.23 & 0.17 & 1.66 & 0.25 & 0.20 & 0.23 & 0.38 & 0.00 & 0.31 \\
    Deepseek-llm-7B-Chat & 0.44 & 0.47 & 1.76 & 1.67 & 0.53 & 0.27 & 1.80 & 0.69 & 1.21 & 1.28 & 2.03 & 0.08 & 1.02 & 1.99 & 2.23 & 0.66 & 3.81 & 0.00 & 2.59 & 1.91 \\
    Seed-Coder-8B-Instruct & 18.34 & 18.91 & 18.00 & 16.19 & 9.40 & 10.07 & 15.91 & 8.78 & 14.07 & 11.99 & 17.39 & 11.34 & 9.26 & 10.83 & 9.58 & 7.11 & 16.02 & 12.39 & 9.29 & 11.67 \\ 
    Llama-3-8B-Instruct & 12.33 & 15.34 & 5.40 & 15.60 & 11.04 & 4.26 & 16.09 & 6.93 & 9.89 & 9.51 & 5.15 & 12.70 & 8.53 & 11.50 & 7.23 & 5.74 & 11.87 & 6.35 & 7.40 & 9.67 \\ 
    Llama-3.1-8B-Instruct & 6.19 & 10.20 & 15.44 & 11.63 & 7.95 & 5.95 & 10.60 & 11.64 & 10.47 & 12.88 & 9.02 & 9.11 & 8.63 & 7.19 & 4.17 & 7.11 & 13.79 & 6.78 & 6.15 & 7.28 \\ 
    Qwen3-8B-nothinking & 11.37 & 12.97 & 13.78 & 12.58 & 8.99 & 6.37 & 18.00 & 8.62 & 17.12 & 13.88 & 8.87 & 16.81 & 17.87 & 12.69 & 12.39 & 16.32 & 15.67 & 13.09 & 13.37 & 16.02 \\ 
    Qwen3-8B-thinking & 30.50 & 35.31 & 35.19 & 34.92 & 26.66 & 31.87 & 33.93 & 31.11 & 36.57 & 26.91 & 32.78 & 27.52 & 17.11 & 33.02 & 6.38 & 37.55 & 38.21 & 36.83 & 27.97 & 37.51 \\ 
    GLM-4-9B-0414-nothinking & 20.10 & 17.02 & 18.55 & 13.43 & 20.23 & 12.22 & 15.56 & 10.97 & 14.93 & 8.09 & 17.57 & 13.26 & 14.61 & 8.79 & 7.00 & 10.82 & 15.45 & 16.94 & 13.44 & 14.25 \\ 
    GLM-4-9B-0414-thinking & 20.40 & 24.90 & 23.86 & 31.73 & 21.47 & 18.93 & 24.03 & 26.68 & 22.28 & 12.62 & 24.84 & 17.54 & 15.29 & 11.78 & 10.05 & 17.94 & 22.79 & 24.36 & 15.54 & 26.78 \\ 
    Phi-4-14B & 27.79 & 30.67 & 24.07 & 36.93 & 23.35 & 29.28 & 30.60 & 28.57 & 30.82 & 21.71 & 30.17 & 27.77 & 17.96 & 25.88 & 14.54 & 26.38 & 28.83 & 28.62 & 32.84 & 30.95 \\ 
    Moonlight-16B-A3B-Instruct & 4.78 & 4.55 & 7.60 & 2.18 & 5.89 & 4.26 & 7.65 & 3.84 & 3.55 & 3.20 & 2.93 & 5.45 & 4.59 & 1.22 & 4.65 & 3.02 & 3.95 & 4.92 & 2.68 & 5.98 \\ 
    ERNIE-4.5-21B-A3B-PT & 25.96 & 25.94 & 22.80 & 31.89 & 16.04 & 23.50 & 31.46 & 20.38 & 17.65 & 23.60 & 16.71 & 21.25 & 11.98 & 19.66 & 13.34 & 21.06 & 24.17 & 19.22 & 23.16 & 27.70 \\ 
    GLM-4-32B & 35.14 & 34.82 & 39.41 & 41.36 & 27.32 & 38.97 & 38.40 & 33.89 & 34.54 & 26.95 & 34.58 & 33.44 & 23.71 & 29.08 & 22.86 & 38.54 & 32.68 & 34.75 & 35.29 & 42.16 \\ 
    Qwen3-32B-nothinking & 18.12 & 21.02 & 17.85 & 16.51 & 22.02 & 11.91 & 21.55 & 14.55 & 18.63 & 22.72 & 17.43 & 19.16 & 16.24 & 23.32 & 14.70 & 25.27 & 23.53 & 24.97 & 16.39 & 14.33 \\ 
    Qwen3-32B-thinking & 47.66 & 47.09 & 48.05 & 52.14 & 41.65 & 43.28 & 43.74 & 37.39 & 39.31 & 45.76 & 45.13 & 42.39 & 30.64 & 37.59 & 17.88 & 49.55 & 45.99 & 55.76 & 38.61 & 42.35 \\ \midrule
\multicolumn{21}{c}{\textbf{Close-source Methods}} \\ \midrule
    GPT-3.5-turbo & 17.02 & 20.05 & 20.03 & 19.17 & 14.62 & 13.98 & 20.26 & 16.53 & 17.98 & 9.66 & 19.70 & 20.49 & 16.63 & 12.07 & 12.16 & 14.01 & 15.82 & 17.96 & 21.58 & 15.20 \\ 
    GPT-4o & 37.49 & 38.32 & 36.46 & 42.00 & 33.55 & 30.81 & 46.27 & 36.31 & 32.76 & 35.13 & 36.50 & 36.07 & 28.70 & 42.73 & 29.07 & 39.34 & 33.86 & 38.31 & 39.76 & 40.77 \\ 
    Gemini-2.0 & 34.30 & 24.48 & 30.76 & 32.02 & 32.07 & 33.71 & 28.24 & 27.62 & 27.05 & 33.75 & 28.84 & 22.35 & 22.92 & 33.77 & 28.05 & 27.97 & 24.91 & 23.74 & 23.11 & 31.10 \\ 
    Gemini-2.5 & 64.13 & 52.14 & 64.69 & 61.28 & 57.65 & 53.92 & 56.84 & 55.01 & 57.85 & 51.57 & 45.55 & 56.85 & 35.04 & 53.90 & 38.28 & 40.51 & 48.27 & 62.07 & 54.41 & 52.37 \\ 
    Deepseek-Chat-V3 & 50.45 & 43.23 & 47.42 & 45.85 & 41.53 & 42.38 & 45.96 & 44.42 & 39.53 & 38.79 & 41.64 & 47.36 & 32.88 & 47.61 & 32.39 & 47.34 & 42.20 & 51.28 & 50.14 & 50.96 \\ \midrule
\multicolumn{21}{c}{\textbf{Open-Source Training Methods}} \\ \midrule
    Llama-3.1-8B-Instruct-SFT & 41.45 & 42.03 & 41.94 & 39.98 & 35.69 & 34.48 & 38.30 & 30.82 & 34.22 & 35.38 & 37.74 & 40.46 & 26.75 & 30.88 & 26.53 & 35.24 & 36.67 & 39.66 & 39.10 & 40.88 \\ 
    Llama-3.1-8B-Instruct-SFT-GRPO & 58.43 & 54.60 & 46.89 & 52.47 & 45.94 & 43.43 & 40.80 & 47.19 & 51.75 & 48.92 & 44.47 & 55.18 & 51.66 & 46.54 & 36.46 & 60.17 & 49.25 & 53.63 & 52.68 & 50.80 \\ 
    
    Qwen3-8B-SFT & 45.36 & 39.48 & 40.77 & 46.45 & 39.79 & 35.69 & 47.72 & 38.67 & 39.53 & 34.84 & 40.54 & 42.44 & 28.31 & 38.71 & 27.43 & 46.25 & 35.56 & 50.07 & 47.08 & 48.54 \\
    Qwen3-8B-SFT-GRPO & 52.67 & 50.61 & 47.72 & 46.62 & 43.87 & 47.68 & 50.40 & 40.54 & 46.62 & 48.78 & 43.80 & 46.23 & 22.15 & 45.45 & 26.67 & 45.38 & 44.98 & 50.84 & 46.31 & 60.57 \\ \midrule
Method & Ga & It & Ja & Jav & Xal & Kn & Kk & Rw & Kon & Ko & Ky & Lao & Lat & Lv & Li & Mk & Mg & Ms & Ml & Mt \\ \midrule
\multicolumn{21}{c}{\textbf{Open-source Methods}} \\ \midrule
    Baichuan2-7B-Chat & 1.73 & 1.51 & 1.11 & 0.14 & 0.13 & 1.45 & 0.32 & 0.89 & 0.22 & 0.12 & 0.21 & 0.28 & 0.59 & 0.00 & 0.99 & 1.20 & 0.00 & 0.74 & 0.00 & 0.18 \\
    Deepseek-llm-7B-Chat & 0.00 & 2.09 & 1.11 & 2.40 & 0.50 & 1.81 & 0.53 & 0.00 & 5.07 & 0.00 & 0.00 & 1.69 & 1.07 & 0.34 & 0.69 & 0.46 & 0.36 & 2.76 & 0.58 & 0.23 \\
    Seed-Coder-8B-Instruct & 7.12 & 20.27 & 11.22 & 12.28 & 6.95 & 6.36 & 13.86 & 12.78 & 10.18 & 9.31 & 8.94 & 13.93 & 8.81 & 16.56 & 8.95 & 12.20 & 3.52 & 15.74 & 5.60 & 8.17 \\ 
    Llama-3-8B-Instruct & 11.04 & 7.14 & 4.81 & 8.89 & 5.81 & 6.65 & 17.00 & 7.63 & 16.95 & 8.07 & 7.08 & 0.57 & 9.69 & 9.02 & 11.23 & 10.74 & 5.60 & 10.92 & 10.80 & 10.07 \\ 
    Llama-3.1-8B-Instruct & 7.93 & 10.89 & 6.44 & 9.79 & 11.03 & 4.30 & 7.83 & 5.81 & 7.79 & 10.63 & 5.20 & 6.11 & 5.61 & 7.71 & 7.25 & 10.80 & 4.83 & 12.45 & 5.12 & 7.79 \\
    Qwen3-8B-nothinking & 13.54 & 14.78 & 11.71 & 11.86 & 12.20 & 11.40 & 12.51 & 10.27 & 21.44 & 13.65 & 7.97 & 12.20 & 9.55 & 11.98 & 12.32 & 16.13 & 9.04 & 18.80 & 11.70 & 13.29 \\  
    Qwen3-8B-thinking & 26.16 & 29.96 & 31.52 & 32.47 & 32.14 & 29.53 & 36.48 & 12.87 & 17.42 & 31.89 & 21.01 & 19.30 & 25.46 & 37.72 & 35.55 & 34.84 & 10.46 & 36.20 & 23.79 & 23.41 \\ 
    GLM-4-9B-0414-nothinking & 10.14 & 15.93 & 11.23 & 10.92 & 10.55 & 6.62 & 17.57 & 6.08 & 7.78 & 23.60 & 6.91 & 4.75 & 6.45 & 10.27 & 11.28 & 10.25 & 7.94 & 17.21 & 8.03 & 10.04 \\ 
    GLM-4-9B-0414-thinking & 20.74 & 18.64 & 26.67 & 21.67 & 17.75 & 9.28 & 22.81 & 6.13 & 7.35 & 22.19 & 13.73 & 6.64 & 12.70 & 21.92 & 21.21 & 19.70 & 5.85 & 31.83 & 10.13 & 19.57 \\ 
    Phi-4-14B & 22.40 & 29.27 & 29.26 & 28.10 & 22.54 & 22.10 & 33.86 & 10.37 & 14.33 & 33.66 & 24.35 & 14.35 & 38.65 & 25.22 & 35.82 & 26.88 & 16.03 & 36.83 & 31.19 & 22.78 \\ 
    Moonlight-16B-A3B-Instruct & 4.90 & 3.60 & 6.50 & 6.80 & 2.54 & 2.55 & 4.41 & 3.68 & 6.55 & 3.77 & 2.69 & 2.34 & 4.45 & 1.40 & 4.77 & 7.17 & 1.12 & 4.29 & 0.46 & 4.52 \\ 
    ERNIE-4.5-21B-A3B-PT & 18.74 & 21.88 & 20.94 & 20.87 & 23.31 & 20.70 & 28.61 & 15.92 & 19.91 & 17.50 & 23.45 & 21.69 & 26.55 & 25.52 & 26.72 & 24.89 & 21.00 & 23.82 & 27.25 & 25.02 \\ 
    GLM-4-32B & 32.72 & 32.53 & 35.50 & 35.94 & 31.75 & 28.66 & 33.25 & 10.96 & 12.99 & 38.96 & 30.73 & 12.70 & 30.50 & 31.15 & 31.19 & 37.12 & 21.14 & 41.19 & 31.35 & 30.53 \\ 
    Qwen3-32B-nothinking & 25.94 & 21.78 & 17.54 & 11.89 & 14.84 & 21.17 & 17.05 & 13.08 & 21.52 & 16.72 & 12.86 & 13.84 & 24.91 & 14.79 & 19.24 & 23.98 & 10.56 & 19.96 & 24.38 & 11.98 \\ 
    Qwen3-32B-thinking & 51.40 & 44.27 & 44.67 & 38.98 & 42.19 & 39.85 & 48.52 & 17.46 & 20.93 & 44.43 & 45.25 & 28.89 & 41.36 & 49.09 & 46.77 & 49.34 & 21.62 & 47.87 & 42.61 & 29.30 \\ \midrule
\multicolumn{21}{c}{\textbf{Close-source Methods}} \\ \midrule
    GPT-3.5-turbo & 13.51 & 16.14 & 18.08 & 16.88 & 12.21 & 13.57 & 18.30 & 12.74 & 16.07 & 18.32 & 11.24 & 7.43 & 17.91 & 22.04 & 16.18 & 19.12 & 14.89 & 22.25 & 18.40 & 15.74 \\ 
    GPT-4o & 44.73 & 37.88 & 39.26 & 31.27 & 35.48 & 42.43 & 41.74 & 20.91 & 38.93 & 37.75 & 34.42 & 15.58 & 33.92 & 33.19 & 43.51 & 38.75 & 30.56 & 36.98 & 39.48 & 39.22 \\ 
    Gemini-2.0 & 31.34 & 27.07 & 31.59 & 29.89 & 32.18 & 36.35 & 40.26 & 25.28 & 32.88 & 21.53 & 24.68 & 21.42 & 29.24 & 31.59 & 29.92 & 31.99 & 24.18 & 32.57 & 37.34 & 34.53 \\ 
    Gemini-2.5 & 60.93 & 52.82 & 51.41 & 56.02 & 48.20 & 43.63 & 60.23 & 33.90 & 39.48 & 54.81 & 47.47 & 42.61 & 53.01 & 52.93 & 55.36 & 51.29 & 57.38 & 53.28 & 55.95 & 47.22 \\ 
    Deepseek-Chat-V3 & 52.58 & 49.42 & 43.40 & 42.80 & 43.84 & 35.31 & 44.37 & 23.30 & 41.95 & 40.96 & 40.74 & 36.48 & 43.48 & 48.54 & 48.06 & 46.49 & 45.43 & 50.73 & 45.96 & 34.56 \\ \midrule
\multicolumn{21}{c}{\textbf{Open-Source Training Methods}} \\ \midrule
    Llama-3.1-8B-Instruct-SFT & 41.80 & 38.50 & 33.06 & 27.05 & 36.37 & 35.46 & 41.67 & 25.93 & 28.36 & 33.67 & 35.83 & 26.48 & 36.94 & 41.05 & 37.92 & 41.05 & 24.43 & 38.28 & 34.12 & 32.31 \\ 
    Llama-3.1-8B-Instruct-SFT-GRPO & 57.36 & 55.27 & 52.42 & 49.09 & 43.74 & 49.40 & 50.75 & 48.95 & 43.00 & 46.28 & 48.20 & 52.80 & 52.13 & 51.84 & 55.67 & 46.19 & 45.64 & 47.93 & 47.92 & 51.18 \\ 
    
    Qwen3-8B-SFT & 45.73 & 38.49 & 36.11 & 37.92 & 45.20 & 38.27 & 47.66 & 21.37 & 33.72 & 39.62 & 36.51 & 30.57 & 42.32 & 44.14 & 47.72 & 40.80 & 25.40 & 41.91 & 38.52 & 36.99 \\
    Qwen3-8B-SFT-GRPO & 47.79 & 47.25 & 46.33 & 45.72 & 43.57 & 42.10 & 51.20 & 35.47 & 32.11 & 48.78 & 41.83 & 31.99 & 41.99 & 52.49 & 49.53 & 49.04 & 38.02 & 48.84 & 47.56 & 43.66 \\ \midrule
Method & Mri & Mr & Mon & Mos & Ne & No & Ps & Fa & Pl & Pt & Pa & Que & Ro & Rom & Ru & San & Sr & Si & Sk & Sl \\ \midrule
\multicolumn{21}{c}{\textbf{Open-source Methods}} \\ \midrule
    Baichuan2-7B-Chat & 1.23 & 0.08 & 0.22 & 0.00 & 0.69 & 0.25 & 1.30 & 0.78 & 0.87 & 0.53 & 0.00 & 0.08 & 0.54 & 1.41 & 1.18 & 0.70 & 0.38 & 0.10 & 0.21 & 1.88 \\
    Deepseek-llm-7B-Chat & 1.99 & 1.48 & 0.04 & 0.38 & 1.38 & 2.63 & 0.58 & 1.51 & 1.33 & 1.30 & 1.00 & 1.11 & 2.92 & 0.30 & 0.20 & 1.63 & 1.02 & 0.00 & 0.79 & 0.57 \\
    Seed-Coder-8B-Instruct & 8.27 & 7.05 & 9.04 & 2.76 & 9.40 & 12.01 & 10.53 & 15.20 & 16.17 & 9.87 & 7.08 & 10.46 & 17.85 & 11.29 & 16.06 & 9.47 & 18.90 & 3.46 & 14.96 & 13.61 \\ 
    Llama-3-8B-Instruct & 9.02 & 5.12 & 9.45 & 6.31 & 5.22 & 11.19 & 13.55 & 12.22 & 11.42 & 11.97 & 8.18 & 6.34 & 15.68 & 7.61 & 9.65 & 8.32 & 16.01 & 5.00 & 8.79 & 12.42 \\ 
    Llama-3.1-8B-Instruct & 8.86 & 7.37 & 9.28 & 8.80 & 4.57 & 12.00 & 8.41 & 13.67 & 14.03 & 8.32 & 6.67 & 5.94 & 15.72 & 6.25 & 8.59 & 10.73 & 7.25 & 2.49 & 12.24 & 5.49 \\ 
    Qwen3-8B-nothinking & 11.89 & 11.81 & 13.22 & 9.33 & 13.58 & 14.24 & 14.13 & 17.46 & 14.38 & 10.77 & 13.51 & 5.46 & 19.13 & 9.88 & 10.59 & 10.48 & 13.05 & 9.39 & 15.90 & 11.30 \\ 
    Qwen3-8B-thinking & 10.31 & 20.93 & 29.48 & 7.08 & 27.73 & 38.00 & 23.74 & 33.84 & 30.85 & 36.49 & 26.40 & 7.51 & 38.37 & 17.72 & 29.36 & 30.10 & 38.07 & 16.77 & 37.24 & 29.16 \\ 
    GLM-4-9B-0414-nothinking & 5.32 & 10.47 & 9.53 & 9.23 & 12.14 & 22.79 & 7.43 & 21.97 & 20.21 & 21.62 & 14.37 & 6.21 & 19.58 & 6.25 & 15.00 & 16.42 & 23.41 & 4.74 & 17.09 & 12.43 \\ 
    GLM-4-9B-0414-thinking & 4.75 & 11.95 & 15.86 & 9.97 & 23.05 & 21.56 & 17.58 & 25.52 & 22.25 & 24.31 & 14.91 & 11.15 & 30.16 & 11.38 & 21.38 & 18.32 & 27.77 & 8.08 & 27.38 & 21.14 \\ 
    Phi-4-14B & 14.83 & 23.14 & 17.35 & 11.71 & 25.51 & 31.88 & 26.86 & 34.62 & 31.96 & 33.94 & 29.14 & 19.98 & 32.33 & 14.05 & 26.42 & 32.34 & 29.39 & 12.32 & 31.83 & 25.07 \\ 
    Moonlight-16B-A3B-Instruct & 3.02 & 1.00 & 5.43 & 3.43 & 3.68 & 7.24 & 5.12 & 6.25 & 8.31 & 6.11 & 1.74 & 6.05 & 7.73 & 1.93 & 3.73 & 6.12 & 6.18 & 2.43 & 5.70 & 4.21 \\ 
    ERNIE-4.5-21B-A3B-PT & 19.39 & 23.96 & 14.44 & 13.51 & 25.18 & 26.33 & 26.46 & 27.69 & 24.26 & 28.57 & 20.32 & 11.38 & 27.03 & 16.00 & 24.48 & 22.77 & 33.10 & 21.43 & 26.00 & 19.68 \\ 
    GLM-4-32B & 17.06 & 26.03 & 20.83 & 11.93 & 36.03 & 36.98 & 29.45 & 38.58 & 32.08 & 38.53 & 31.97 & 14.34 & 31.35 & 30.18 & 32.84 & 28.48 & 35.75 & 25.66 & 37.26 & 27.99 \\ 
    Qwen3-32B-nothinking & 21.09 & 22.85 & 21.28 & 9.45 & 27.72 & 23.20 & 21.19 & 26.90 & 22.01 & 16.36 & 21.70 & 9.40 & 23.88 & 18.20 & 21.52 & 24.34 & 25.66 & 22.74 & 19.41 & 17.29 \\ 
    Qwen3-32B-thinking & 36.78 & 42.34 & 39.86 & 12.19 & 47.94 & 45.88 & 32.47 & 40.59 & 37.98 & 44.08 & 39.93 & 17.40 & 43.50 & 24.58 & 43.27 & 43.83 & 44.02 & 38.74 & 47.30 & 49.72 \\ \midrule
\multicolumn{21}{c}{\textbf{Close-source Methods}} \\ \midrule
    GPT-3.5-turbo & 16.23 & 10.82 & 17.48 & 6.87 & 13.52 & 26.06 & 14.22 & 22.67 & 21.59 & 15.68 & 13.94 & 11.44 & 26.24 & 11.70 & 16.57 & 12.74 & 19.80 & 8.81 & 18.43 & 17.64 \\ 
    GPT-4o & 36.17 & 28.16 & 39.80 & 17.38 & 40.14 & 37.30 & 39.82 & 42.93 & 37.98 & 35.10 & 41.40 & 26.90 & 39.25 & 20.70 & 35.11 & 35.55 & 36.66 & 40.93 & 39.56 & 34.28 \\ 
    Gemini-2.0 & 35.78 & 21.97 & 26.80 & 23.04 & 34.60 & 26.85 & 40.73 & 25.06 & 20.93 & 26.64 & 27.84 & 17.26 & 39.45 & 27.79 & 26.71 & 29.41 & 33.22 & 34.84 & 29.75 & 23.37 \\ 
    Gemini-2.5 & 49.45 & 47.82 & 45.59 & 18.32 & 48.39 & 56.15 & 52.71 & 50.19 & 52.88 & 55.05 & 55.55 & 34.62 & 57.23 & 49.28 & 48.12 & 46.44 & 47.07 & 50.11 & 58.73 & 52.07 \\ 
    Deepseek-Chat-V3 & 38.88 & 37.28 & 43.56 & 14.49 & 43.79 & 42.79 & 36.30 & 50.04 & 45.26 & 45.84 & 39.91 & 22.24 & 40.51 & 26.15 & 45.94 & 42.95 & 49.79 & 40.47 & 49.86 & 42.59 \\ \midrule
\multicolumn{21}{c}{\textbf{Open-Source Training Methods}} \\ \midrule
    Llama-3.1-8B-Instruct-SFT & 36.84 & 32.20 & 28.98 & 15.77 & 33.85 & 34.69 & 34.92 & 37.62 & 37.00 & 43.79 & 39.74 & 19.61 & 38.95 & 24.81 & 40.65 & 38.48 & 40.48 & 31.35 & 42.00 & 40.79 \\ 
    Llama-3.1-8B-Instruct-SFT-GRPO & 50.39 & 48.06 & 47.78 & 41.80 & 52.08 & 58.64 & 48.89 & 53.42 & 53.13 & 47.95 & 41.34 & 47.18 & 44.29 & 51.72 & 47.41 & 55.62 & 48.92 & 45.11 & 51.05 & 46.71 \\ 
    
    Qwen3-8B-SFT & 31.58 & 34.86 & 36.97 & 13.67 & 35.06 & 38.60 & 37.25 & 45.86 & 38.43 & 42.26 & 45.97 & 27.96 & 42.70 & 36.83 & 37.72 & 37.89 & 43.93 & 32.39 & 47.25 & 40.83 \\
    Qwen3-8B-SFT-GRPO & 43.52 & 44.26 & 47.66 & 13.61 & 42.46 & 46.65 & 46.00 & 51.86 & 40.38 & 45.06 & 45.97 & 27.40 & 50.15 & 35.10 & 42.99 & 42.45 & 45.48 & 47.53 & 55.81 & 44.72 \\ \midrule
Method & Som & Es & Swa & Sv & Tg & Ta & Te & Th & Tr & Uk & Ur & Ug & Uz & Vi & Xh & Yor & Zu & & & \\  \midrule
\multicolumn{21}{c}{\textbf{Open-source Methods}} \\ \midrule
    Baichuan2-7B-Chat & 2.63 & 0.18 & 0.25 & 0.28 & 1.02 & 0.13 & 1.92 & 0.22 & 0.22 & 2.16 & 1.39 & 1.15 & 0.30 & 0.00 & 2.00 & 0.00 & 0.00 & & & \\
    Deepseek-llm-7B-Chat & 0.53 & 2.25 & 2.87 & 2.54 & 0.14 & 0.00 & 0.00 & 3.73 & 1.87 & 0.70 & 3.43 & 0.00 & 0.30 & 0.80 & 4.38 & 5.56 & 0.00 & & & \\
    Seed-Coder-8B-Instruct & 8.21 & 24.90 & 11.27 & 11.48 & 9.06 & 10.01 & 13.11 & 9.57 & 7.83 & 17.26 & 17.86 & 9.92 & 8.78 & 6.84 & 10.32 & 16.67 & 2.04 & & & \\ 
    Llama-3-8B-Instruct & 8.17 & 9.08 & 8.90 & 14.93 & 11.45 & 9.80 & 9.45 & 10.72 & 8.32 & 13.61 & 17.35 & 9.78 & 11.07 & 11.93 & 8.28 & 12.96 & 3.87 & & & \\ 
    Llama-3.1-8B-Instruct & 10.97 & 14.57 & 12.92 & 15.55 & 8.21 & 7.93 & 8.34 & 6.89 & 5.03 & 8.40 & 13.96 & 10.13 & 10.90 & 7.98 & 4.17 & 11.11 & 1.95 & & & \\ 
    Qwen3-8B-nothinking & 10.88 & 13.32 & 11.36 & 12.60 & 13.69 & 17.60 & 13.02 & 11.39 & 13.08 & 12.63 & 25.70 & 11.56 & 7.29 & 13.97 & 6.51 & 25.97 & 3.21 & & & \\ 
    Qwen3-8B-thinking & 12.74 & 39.66 & 21.42 & 30.51 & 18.90 & 31.33 & 26.79 & 27.20 & 37.69 & 32.37 & 37.63 & 17.19 & 34.94 & 28.78 & 11.93 & 18.52 & 13.61 & & & \\ 
    GLM-4-9B-0414-nothinking & 8.60 & 24.63 & 6.57 & 19.57 & 13.18 & 16.02 & 11.21 & 8.33 & 16.63 & 19.86 & 13.26 & 5.62 & 9.24 & 19.08 & 12.30 & 5.56 & 4.28 & & & \\ 
    GLM-4-9B-0414-thinking & 9.01 & 25.71 & 14.89 & 25.33 & 14.28 & 13.65 & 10.75 & 14.40 & 30.25 & 26.42 & 29.58 & 7.04 & 22.07 & 27.80 & 13.33 & 18.52 & 0.92 & & & \\ 
    Phi-4-14B & 12.74 & 29.91 & 27.08 & 32.23 & 20.64 & 26.41 & 37.71 & 24.55 & 29.62 & 28.89 & 34.71 & 19.28 & 24.29 & 31.12 & 16.07 & 21.60 & 10.06 & & & \\ 
    Moonlight-16B-A3B-Instruct & 3.48 & 7.13 & 3.00 & 9.23 & 6.12 & 4.61 & 0.31 & 0.61 & 3.42 & 7.83 & 1.56 & 2.97 & 1.52 & 2.25 & 4.83 & 2.69 & 3.92 & & & \\ 
    ERNIE-4.5-21B-A3B-PT & 19.55 & 21.32 & 22.23 & 28.17 & 20.43 & 29.68 & 19.36 & 25.11 & 20.30 & 27.49 & 30.34 & 19.23 & 19.18 & 20.34 & 15.55 & 16.39 & 14.21 & & & \\ 
    GLM-4-32B & 17.86 & 44.97 & 35.40 & 32.65 & 16.24 & 31.10 & 22.89 & 23.68 & 28.79 & 38.45 & 35.96 & 22.90 & 32.15 & 33.84 & 18.57 & 24.28 & 17.00 & & & \\ 
    Qwen3-32B-nothinking & 18.83 & 19.79 & 17.60 & 22.29 & 13.03 & 31.49 & 27.01 & 18.97 & 20.22 & 25.11 & 24.89 & 16.58 & 17.37 & 30.76 & 10.28 & 30.56 & 5.28 & & & \\ 
    Qwen3-32B-thinking & 22.58 & 51.52 & 38.12 & 41.77 & 37.22 & 46.06 & 38.72 & 39.23 & 41.72 & 42.60 & 45.33 & 32.80 & 45.91 & 40.35 & 20.90 & 63.89 & 23.48 & & & \\ \midrule
\multicolumn{21}{c}{\textbf{Close-source Methods}} \\ \midrule
    GPT-3.5-turbo & 12.74 & 21.14 & 19.02 & 24.48 & 11.03 & 13.67 & 15.22 & 15.20 & 19.87 & 21.32 & 24.89 & 9.97 & 14.89 & 21.85 & 11.57 & 24.72 & 12.83 & & & \\ 
    GPT-4o & 36.46 & 34.87 & 36.95 & 33.98 & 37.52 & 44.01 & 36.62 & 32.82 & 35.70 & 41.20 & 42.86 & 27.95 & 42.06 & 40.78 & 33.89 & 35.42 & 27.91 & & & \\ 
    Gemini-2.0 & 33.62 & 31.60 & 28.56 & 23.85 & 33.56 & 45.47 & 32.63 & 25.03 & 24.57 & 33.07 & 39.81 & 28.47 & 26.66 & 32.27 & 35.42 & 33.21 & 24.98 & & & \\ 
    Gemini-2.5 & 52.84 & 59.42 & 43.79 & 52.67 & 49.17 & 61.05 & 51.77 & 47.56 & 50.10 & 54.25 & 55.51 & 44.65 & 54.36 & 56.64 & 42.13 & 49.72 & 48.16 \\ 
    Deepseek-Chat-V3 & 39.37 & 50.02 & 40.43 & 41.61 & 37.44 & 45.68 & 46.89 & 36.76 & 46.25 & 43.47 & 49.87 & 39.21 & 46.43 & 46.35 & 33.25 & 39.76 & 23.93 & & & \\ \midrule
\multicolumn{21}{c}{\textbf{Open-Source Training Methods}} \\ \midrule
    Llama-3.1-8B-Instruct-SFT & 32.95 & 39.15 & 28.39 & 35.83 & 26.71 & 41.01 & 33.51 & 30.11 & 29.54 & 32.21 & 43.97 & 33.52 & 38.54 & 42.82 & 25.26 & 58.74 & 29.87 & & & \\ 
    Llama-3.1-8B-Instruct-SFT-GRPO & 49.48 & 50.12 & 37.27 & 44.07 & 49.39 & 45.06 & 41.06 & 41.49 & 55.15 & 44.70 & 59.93 & 48.12 & 52.44 & 44.33 & 49.41 & 49.76 & 44.85 & & & \\ 
    
    Qwen3-8B-SFT & 37.76 & 43.82 & 35.98 & 38.00 & 30.56 & 51.78 & 46.06 & 35.62 & 39.18 & 43.64 & 46.47 & 38.56 & 41.52 & 46.25 & 30.21 & 36.33 & 23.49 &  & & \\
    Qwen3-8B-SFT-GRPO & 34.29 & 48.79 & 45.46 & 47.84 & 37.88 & 54.48 & 53.55 & 39.39 & 45.90 & 45.04 & 49.95 & 42.80 & 51.99 & 50.39 & 26.51 & 34.71 & 29.30 &  & & \\
  
\bottomrule
\end{longtable}
}
}
\twocolumn 

\clearpage
\end{document}